\documentclass[10pt,twocolumn,letterpaper]{article}

\usepackage{cvpr}
\usepackage{bm}
\usepackage{times}
\usepackage{epsfig}
\usepackage{graphicx}
\usepackage{amsmath}
\usepackage{amssymb}
\usepackage{subfigure}
\usepackage{multirow}
\usepackage{threeparttable}
\usepackage{arydshln}
\usepackage{color}

\usepackage[pagebackref=true,breaklinks=true,letterpaper=true,colorlinks,bookmarks=false]{hyperref}

\cvprfinalcopy 

\def\cvprPaperID{****} 

\ifcvprfinal\fi
\begin{document}

\title{Wavelet Integrated CNNs for Noise-Robust Image Classification}

\author{
Qiufu Li$^{1,2}$\quad Linlin Shen
\thanks{Corresponding Author: Linlin Shen.}
\ $^{1,2}$\quad Sheng Guo$^3$\quad Zhihui Lai$^{1,2}$\\
$^1$Computer Vision Institute, School of Computer Science \& Software Engineering, Shenzhen University\\
$^2$Shenzhen Institute of Artificial Intelligence \& Robotics for Society\quad
$^3$Malong Technologies \\
{\tt\small \{liqiufu,llshen\}@szu.edu.cn,sheng@malong.com,lai\_zhi\_hui@163.com}
}

\maketitle
\thispagestyle{empty}

\begin{abstract}
   Convolutional Neural Networks (CNNs) are generally prone to noise interruptions,
   i.e., small image noise can cause drastic changes in the output.
   To suppress the noise effect to the final predication,
   we enhance CNNs by replacing max-pooling, strided-convolution, and average-pooling with Discrete Wavelet Transform (DWT).
   We present general DWT and Inverse DWT (IDWT) layers applicable to various wavelets like Haar, Daubechies, and Cohen, etc.,
   and design wavelet integrated CNNs (WaveCNets) using these layers for image classification.
   In WaveCNets, feature maps are decomposed into the low-frequency and high-frequency components during the down-sampling.
   The low-frequency component stores main information including the basic object structures,
   which is transmitted into the subsequent layers to extract robust high-level features.
   The high-frequency components, containing most of the data noise,
   are dropped during inference to improve the noise-robustness of the WaveCNets.
   Our experimental results on ImageNet and ImageNet-C (the noisy version of ImageNet)
   show that WaveCNets, the wavelet integrated versions of VGG, ResNets, and DenseNet,
   achieve higher accuracy and better noise-robustness than their vanilla versions.
   The code of our DWT/IDWT layer and different WaveCNets are available at {\href{https://github.com/LiQiufu/WaveCNet}{https://github.com/LiQiufu/WaveCNet}}.
\end{abstract}

\section{Introduction}
\begin{figure}[!tbp]
	\centering
	\includegraphics*[scale=0.625, viewport=78 536 451 772]{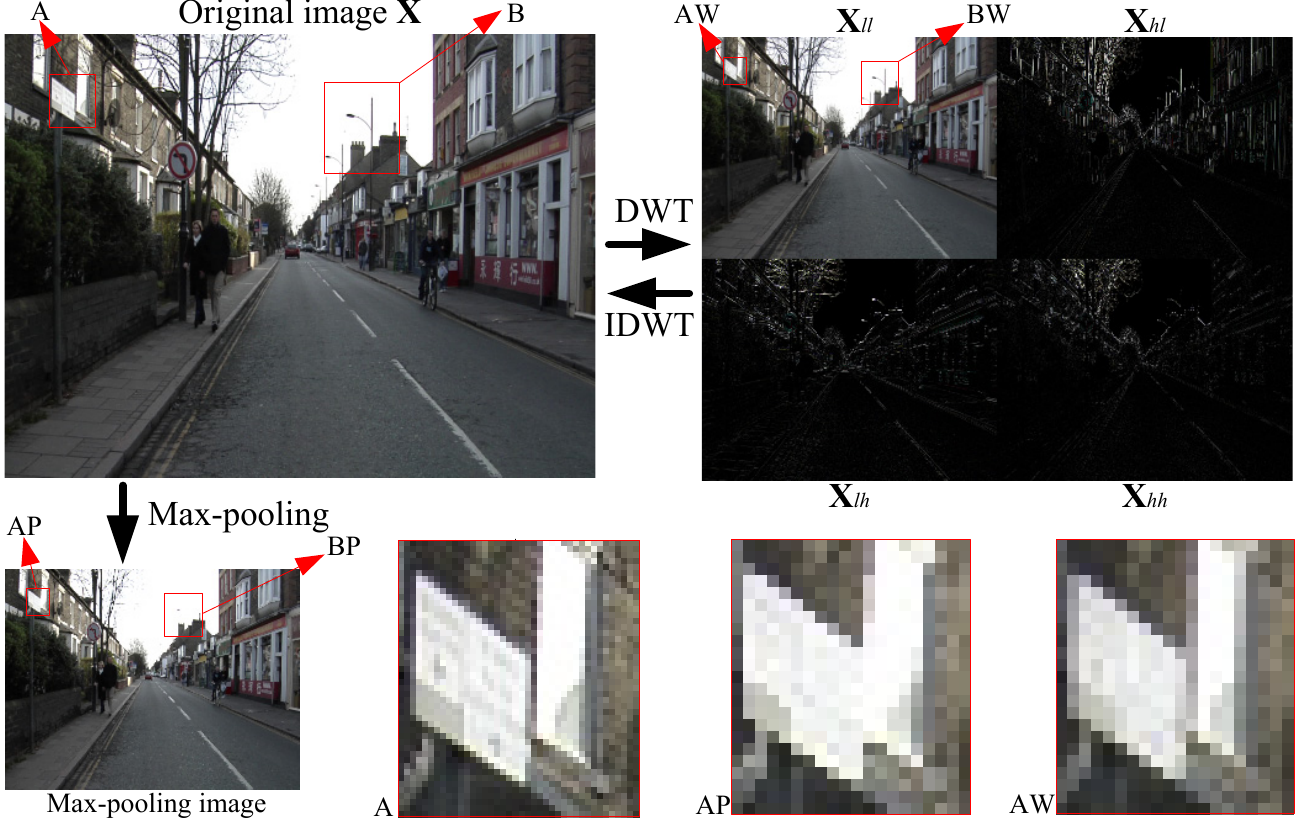}
	\caption{Comparison of max-pooling and wavelet transforms.
		Max-pooling is a commonly used down-sampling operation in the deep networks,
		which could easily breaks the basic object structures.
		Discrete Wavelet Transform (DWT) decomposes an image $\textbf{X}$ into its low-frequency component $\textbf{X}_{ll}$
		and high-frequency components $\textbf{X}_{lh}, \textbf{X}_{hl}, \textbf{X}_{hh}$.
		While $\textbf{X}_{lh}, \textbf{X}_{hl}, \textbf{X}_{hh}$ represent image details including most of the noise,
		$\textbf{X}_{ll}$ is a low resolution version of the data, where the basic object structures are represented.
		In the figures, the window boundary in area A (AP) and the poles in area B (BP) are broken by max-pooling,
		while the principal features of these objects are kept in the DWT output (AW and BW).
	}\label{fig_DWT}
\end{figure}
Drastic changes due to small variations of the input
can emerge in the output of a well-trained convolutional neural network (CNN) for image classification
\cite{goodfellow2014explaining, xie2019feature, geirhos2018imagenet}.
Particularly, the CNN is associated with weak noise-robustness \cite{hendrycks2019benchmarking}.
Random noise of data is mostly high-frequency components.
In the field of signal processing,
transforming the data into different frequency intervals,
and denoising the components in the high-frequency intervals,
is an effective way to denoise it \cite{donoho1995noising,donoho1994ideal}.
The transformation,
such as Discrete Wavelet Transform (DWT) \cite{mallat1989theory},
consists of filtering and down-sampling.
However, the commonly used CNN architectures (VGG, ResNets, and DenseNet, etc.)
do not perform filtering before the feature map down-sampling.

Without the filtering, down-sampling may result in the aliasing among low-frequency and high-frequency components
\cite{nyquist1928certain,zhang2019making}.
In particular, noise in the high-frequency components could be down-sampled into the following feature maps,
and degrade the noise-robustness of the CNNs.
Meanwhile, the basic object structures presented in the low-frequency component could be broken, as Fig. \ref{fig_DWT} shows.

In this paper, to suppress the noise effect to the final predication and increase the classification accuracy,
we integrate wavelet into commonly used CNN architectures.
We firstly transform DWT and Inverse DWT (IDWT) as general network layers in PyTorch \cite{paszke2017automatic}.
Then, we design wavelet integrated convolutional network (WaveCNet),
by replacing the commonly used down-sampling with DWT.
During down-sampling, WaveCNet eliminates the high-frequency components of the feature maps
to increase the noise-robustness of the CNNs,
and then extracts high-level features from the low-frequency component for better classification accuracy.
Using ImageNet \cite{deng2009imagenet} and ImageNet-C \cite{hendrycks2019benchmarking},
we evaluate WaveCNets in terms of classification accuracy and noise-robustness, when various wavelets and various CNN architectures are used.
At last, we explore the application of DWT/IDWT layer in image segmentation.
In summary:
\begin{enumerate}
\setlength{\topsep}{-2ex}
\setlength{\itemsep}{-0.5ex}
\item We present general DWT/IDWT layer applicable to various wavelets, which could be used to design end-to-end wavelet integrated deep networks.
\item We design WaveCNets by replacing existing down-sampling operations with DWT
	  to improve the classification accuracy and noise-robustness of CNNs.
\item We evaluate WaveCNets on ImageNet,
	  and achieve increased accuracy and better noise-robustness.
\item The proposed DWT/IDWT layer is further integrated into SegNet \cite{badrinarayanan2017segnet}
	  to improve the segmentation performance of encoder-decoder networks.
\end{enumerate}

\section{Related works}
\subsection{Noise-robustness}
When the input image is changed, the output of CNN can be significantly different,
regardless of whether the change can be easily perceived by human or not
\cite{goodfellow2014explaining, geirhos2018imagenet, liao2018defense, xie2019feature}.
While the changes may result from various factors,
such as shift \cite{zhang2019making,mairal2014convolutional}, rotation \cite{bruna2013invariant_scatnet},
noise \cite{xie2019feature}, blur \cite{hendrycks2019benchmarking}, manual attack \cite{goodfellow2014explaining}, etc.,
we focus on the robustness of CNNs to the common noise.
A high-level representation guided denoiser is designed in \cite{liao2018defense}
to denoise the contaminated image before inputting it into the CNN,
which may complicate the whole deep network structure.
In \cite{xie2019feature}, the authors propose denoising block for CNNs to denoise the feature map
and suppress the effect of noise on the final prediction.
However, the authors design their denoising block using the spacial filtering,
such as Gaussian filtering, mean filtering, and median filtering, etc.,
which do denoising in the whole frequency domain
and may break basic object structure contained in the low-frequency component.
Therefore, their denoising block requires a residual structure for the CNN to converge.
Recently, a benchmark evaluating CNN performance on noisy images is proposed in \cite{hendrycks2019benchmarking}.
Our WaveCNets will be evaluated using this benchmark.

The recent studies show that ImageNet-trained CNNs prefer to extract features from object textures
sensitive to noise \cite{brendel2019approximating, geirhos2018imagenet}.
Stylized ImageNet \cite{geirhos2018imagenet} is proposed via stylizing ImageNet images with style transfer
to enable the CNNs to extract more robust features from object structures.
The noise could be enlarged as the feature maps flow through layers in the CNNs \cite{liao2018defense, xie2019feature},
resulting in the final wrong predictions.
These issues may be related to the down-sampling operations ignoring the classic sampling theorem.

\subsection{Down-sampling}
For local connectivity and weight sharing, researchers introduce into deep networks various down-sampling operations,
such as max-pooling, average-pooling, mixed-pooling, stochastic pooling, and strided-convolution, etc.
While max-pooling and average-pooling are simple and effective,
they can erase or dilute details from images \cite{yu2014mixed,zeiler2013stochastic}.
Although mixed-pooling \cite{yu2014mixed} and stochastic pooling \cite{zeiler2013stochastic} are introduced to address these issues,
max-pooling, average-pooling, and strided-convolution are still the most widely used operations in CNNs
\cite{he2016deep,huang2017densely,sandler2018mobilenetv2,simonyan2014very}.

These down-sampling operations usually ignore the classic sampling theorem \cite{azulay2018deep,zhang2019making},
which could break object structures and accumulate noise.
Fig. \ref{fig_DWT} shows a max-pooling example.
Anti-aliased CNNs \cite{zhang2019making} integrate the classic anti-aliasing filtering with the down-sampling.
The author is surprised at the increased classification accuracy and better noise-robustness.
Compared to the anti-aliased CNNs, our WaveCNets are significantly different in two aspects:
(1) While Max operation is still used in anti-aliased CNNs, WaveCNets do not require such operation.
(2) The low-pass filters used in anti-aliased CNNs are empirically designed based on the row vectors of Pascal's triangle,
which is ad hoc and no theoretical justifications are given.
As no up-sampling operation, i.e., reconstruction, of the low-pass filter is available,
the anti-aliased U-Net \cite{zhang2019making} has to apply the same filtering after normal up-sampling to achieve the anti-aliasing effect.
In comparison, our WaveCNets are justified by the well defined wavelet theory \cite{daubechies1992ten,mallat1989theory}.
Both down-sampling and up-sampling can be replaced by DWT and IDWT, respectively.

In deep networks for image-to-image translation tasks, the up-sampling operations,
such as transposed convolution in U-Net \cite{ronneberger2015u_net} and max-unpooling in SegNet \cite{badrinarayanan2017segnet},
are widely applied to upgrade the feature map resolution.
Due to the absence of the strict mathematical terms,
these up-sampling operations can not precisely recover the original data.
They do not perform well in the restoration of image details.

\subsection{Wavelets}
Wavelets \cite{daubechies1992ten,mallat1989theory} are powerful time-frequency analysis tools,
which have wide applications in signal processing.
While Discrete Wavelet Transform (DWT) decompose a data into various components in different frequency intervals,
Inverse DWT (IDWT) could reconstruct the data using the DWT output.
DWT could be applied for anti-aliasing in signal processing, and we will explore its application in deep networks.
IDWT could be used for detail restoration in image-to-image tasks.

Wavelet has been combined with neural network for function approximation \cite{zhang1992wavelet},
signal representation and classification \cite{szu1992neural}.
In these early works, the authors apply shallow networks to search the optimal wavelet in wavelet parameter domain.
Recently, this method is utilized with deeper network for image classification,
but the network is difficult to train because of the significant amount of computational cost \cite{de2019multi}.
ScatNet \cite{bruna2013invariant_scatnet} cascades wavelet transform with nonlinear modulus and average pooling,
to extract a translation invariant feature robust to deformations and preserve high-frequency information for image classification.
The authors introduce ScatNet when they explore from mathematical and algorithmic perspective how to design the optimal deep network.
Compared with the CNNs of the same period,
ScatNet gets better performance on the handwritten digit recognition and texture discrimination tasks.
However, due to the strict mathematical assumptions, ScatNet can not be easily transferred to other tasks.

In deep learning, wavelets commonly play the roles of image preprocessing or postprocessing
\cite{huang2017wavelet, liu2018attribute, savareh2019wavelet, yuan2019waveletfcnn}.
Meanwhile, researchers try to introduce wavelet transforms into the design of deep networks in various tasks
\cite{liu2018multi,Williams2018Wavelet,duan2017sar,yoo2019photorealistic}, by taking wavelet transforms as sampling operations.
Multi-level Wavelet CNN (MWCNN) proposed in \cite{liu2018multi} integrates Wavelet Package Transform (WPT)
into the deep network for image restoration.
MWCNN concatenates the low-frequency and high-frequency components of the input feature map,
and processes them in a unified way,
while the data distribution in these components significantly differs from each other.
Convolutional-Wavelet Neural Network (CWNN) proposed in \cite{duan2017sar}
applies dual-tree complex wavelet transform (DT-CWT)
to suppress the noise and keep the structures for extracting robust features from SAR images.
The architecture of CWNN contains only two convolution layers.
While DT-CWT is redundant,
CWNN takes as its down-sampling output the average value of the multiple components output from DT-CWT.
Wavelet pooling proposed in \cite{Williams2018Wavelet} is designed using a two-level DWT.
Its back-propagation performs a one-level DWT and a two-level IDWT, which does not follow the mathematical principle of gradient.
The authors test their method on various dataset
(MNIST \cite{lecun1998gradient}, CIFAR-10 \cite{krizhevsky2009learning}, SHVN \cite{netzer2011reading},
and KDEF \cite{lundqvist1998karolinska}).
However, their network architectures contain only four or five convolutional layers.
The authors do not study systematically the potential of the method on standard image dataset like ImageNet \cite{deng2009imagenet}.
Recently, the application of wavelet transform in image style transfer is studied in \cite{yoo2019photorealistic}.
In above works, the authors evaluate their methods with only one or two wavelets,
due to the absence of the general wavelet transform layers.

\section{Our method}
Our method is trying to apply wavelet transforms to improve the down-sampling operations in deep networks.
We firstly design the general DWT and IDWT layers.

\subsection{DWT and IDWT layers}
The key issues in designs of DWT and IDWT layers are the data forward and backward propagations.
Although the following analysis is for orthogonal wavelet and 1D signal,
it can be generalized to other wavelets and 2D/3D signal with only slight changes.

\textbf{Forward propagation}\quad
For a 1D signal $\textbf{s} = \{s_j\}_{j\in\mathbb{Z}}$,
DWT decomposes it into its low-frequency component $\textbf{s}_1 = \{s_{1k}\}_{k\in\mathbb{Z}}$
and high-frequency component $\textbf{d}_1 = \{d_{1k}\}_{k\in\mathbb{Z}}$,
where
\begin{equation}\label{eq_DWT}
\left\{
\begin{array}{l}
s_{1k} = \sum_j l_{j-2k} s_j, \\
d_{1k} = \sum_j h_{j-2k} s_j,
\end{array}
\right.
\end{equation}
and $\textbf{l} = \{l_k\}_{k\in\mathbb{Z}}, \textbf{h} = \{h_k\}_{k\in\mathbb{Z}}$ are the low-pass
and high-pass filters of an orthogonal wavelet.
According to Eq. (\ref{eq_DWT}), DWT consists of filtering and down-sampling.

Using IDWT, one can reconstruct $\textbf{s}$ from $\textbf{s}_1, \textbf{d}_1$, where
\begin{equation}\label{eq_IDWT}
s_j = \sum_k\left(l_{j-2k} s_{1k} + h_{j-2k} d_{1k}\right).
\end{equation}
In expressions with matrices and vectors, Eq. (\ref{eq_DWT}) and Eq. (\ref{eq_IDWT}) can be rewritten as
\begin{align}
\label{eq_DWT_M_s}
&\textbf{s}_1 = \textbf{L}\textbf{s},\quad\textbf{d}_1 = \textbf{H}\textbf{s},\\
\label{eq_IDWT_M}
&\textbf{s} = \textbf{L}^T \textbf{s}_1 + \textbf{H}^T \textbf{d}_1,
\end{align}
where
\begin{align}
\textbf{L} &=
\left(
\begin{array}{ccccccc}
	\cdots & \cdots & \cdots &        &        &        &                \\
	\cdots & l_{-1} &  l_0   &  l_1   & \cdots &        &                \\
	       &        & \cdots & l_{-1} &  l_0   &  l_1   & \cdots         \\
	       &        &        &        &        & \cdots & \cdots
\end{array}
\right),\\
\textbf{H} &=
\left(
\begin{array}{ccccccc}
	\cdots & \cdots & \cdots &        &        &        &          \\
	\cdots & h_{-1} &  h_0   &  h_1   & \cdots &        &          \\
	       &        & \cdots & h_{-1} &  h_0   &  h_1   & \cdots   \\
	       &        &        &        &        & \cdots & \cdots
\end{array}
\right).
\end{align}

For 2D signal $\textbf{X}$, the DWT usually do 1D DWT
on its every row and column, i.e.,
\begin{align}
\label{eq_DWT_2D_M_ll}
\textbf{X}_{ll} &= \textbf{L} \textbf{X} \textbf{L}^T, \\
\label{eq_DWT_2D_M_lh}
\textbf{X}_{lh} &= \textbf{H} \textbf{X} \textbf{L}^T, \\
\label{eq_DWT_2D_M_hl}
\textbf{X}_{hl} &= \textbf{L} \textbf{X} \textbf{H}^T, \\
\label{eq_DWT_2D_M_hh}
\textbf{X}_{hh} &= \textbf{H} \textbf{X} \textbf{H}^T,
\end{align}
and the corresponding IDWT is implemented with
\begin{equation}\label{eq_IDWT_2D_M}
\textbf{X} = \textbf{L}^T \textbf{X}_{ll} \textbf{L}
+ \textbf{H}^T \textbf{X}_{lh} \textbf{L}
+ \textbf{L}^T \textbf{X}_{hl} \textbf{H}
+ \textbf{H}^T \textbf{X}_{hh} \textbf{H}.
\end{equation}

\textbf{Backward propagation}\quad
For the backward propagation of DWT, we start from Eq. (\ref{eq_DWT_M_s}) and differentiate it,
\begin{equation}\label{eq_DWT_bp}
\frac{\partial \textbf{s}_1}{\partial\textbf{s}} = \textbf{L}^T,\quad \frac{\partial \textbf{d}_1}{\partial\textbf{s}} = \textbf{H}^T.
\end{equation}
Similarly, for the back propagation of the 1D IDWT, differentiate Eq. (\ref{eq_IDWT_M}),
\begin{equation}\label{eq_IDWT_bp}
\frac{\partial{\textbf{s}}}{\partial{\textbf{s}}_1} = \textbf{L},\quad \frac{\partial{\textbf{s}}}{\partial{\textbf{d}}_1} = \textbf{H}.
\end{equation}

The forward and backward propagations of 2D/3D DWT and IDWT are slightly more complicated, but similar to that of 1D DWT and IDWT.
In practice, we choose the wavelets with finite filters, for example,
Haar wavelet with $\textbf{l} = \frac{1}{\sqrt{2}}\{1,1\}$ and $\textbf{h} = \frac{1}{\sqrt{2}}\{1,-1\}$.
For finite signal $\textbf{s} \in \mathbb{R}^N$ and  $\textbf{X} \in \mathbb{R}^{N\times N}$,
the $\textbf{L}, \textbf{H}$ are truncated to be the size of $\lfloor\frac{N}{2}\rfloor\times N$.
We transform 1D/2D/3D DWT and IDWT as network layers in PyTorch.
In the layers, we do DWT and IDWT channel by channel for multi-channel data.

\subsection{WaveCNets}
\begin{figure}[bpt]
	\centering
	\subfigure[The general denoising approach using wavelet.]{
		\label{fig_denoise_a}
		\includegraphics*[scale=0.7, viewport=265 308 585 400]{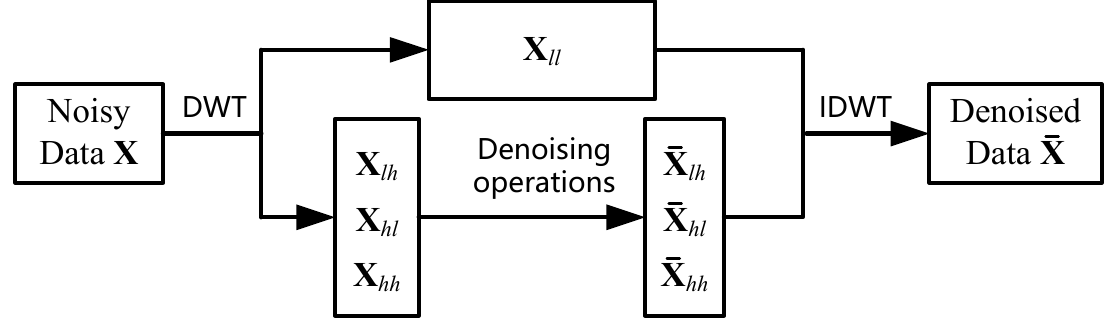}
	}\\
	\subfigure[The simplest wavelet based ``denoising'' method, $\text{DWT}_{ll}$.]{
		\label{fig_denoise_b}
		\includegraphics*[scale=0.7, viewport=265 228 585 258]{figures/Visio-down_sampling_of_dwt.pdf}
	}
	\caption{The general denoising approach based on wavelet transforms and the one used in WaveCNet.}
	\label{fig_dual_structures}
\end{figure}

Given a noisy 2D data $\textbf{X}$, the random noise mostly show up in its high-frequency components.
Therefore, as Fig. \ref{fig_denoise_a} shows, the general wavelet based denoising \cite{donoho1995noising,donoho1994ideal}
consists of three steps:
(1) decompose the noisy data $\textbf{X}$ using DWT into low-frequency component $\textbf{X}_{ll}$
and high-frequency components $\textbf{X}_{lh}, \textbf{X}_{hl}, \textbf{X}_{hh}$,
(2) filter the high-frequency components,
(3) reconstruct the data with the processed components using IDWT.

\begin{figure}[t]
	\centering
	\includegraphics*[scale=0.85, viewport=25 698 295 785]{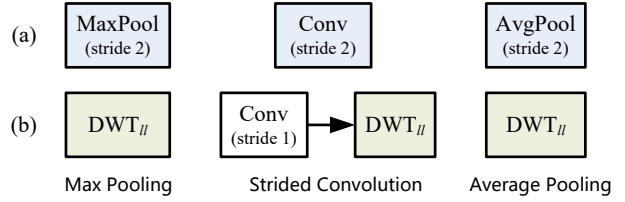}
	\caption{(a) Baseline, the down-sampling operations in deep networks.
	(b) Wavelet integrated down-sampling in WaveCNets.
	}\label{fig_down_sampling_of_dwt}
\end{figure}

In this paper, we choose the simplest wavelet based ``denoising'', i.e., dropping the high-frequency components,
as Fig. \ref{fig_denoise_b} shows.
$\text{DWT}_{ll}$ denotes the transform mapping the feature maps to the low-frequency component.
We design WaveCNets by replacing the commonly used down-sampling with $\text{DWT}_{ll}$.
As Fig. \ref{fig_down_sampling_of_dwt} shows, in WaveCNets, max-pooling and average-pooling are directly replaced by $\text{DWT}_{ll}$,
while strided-convolution is upgrated using convolution with stride of $1$ followed by $\text{DWT}_{ll}$,
i.e.,
\begin{align}
\label{eq_maxpool_up}
\text{MaxPool}_{s = 2} \rightarrow &\ \text{DWT}_{ll},\\
\label{eq_conv_up}
\text{Conv}_{s = 2} \rightarrow &\ \text{DWT}_{ll} \circ \text{Conv}_{s=1},\\
\label{eq_avgpool_up}
\text{AvgPool}_{s = 2} \rightarrow &\ \text{DWT}_{ll},
\end{align}
where ``$\text{MaxPool}_s$'', ``$\text{Conv}_s$'' and ``$\text{AvgPool}_s$'' denote the max-pooling,
strided-convolution, and average-pooling with stride $s$, respectively.

While $\text{DWT}_{ll}$ halves the size of the feature maps,
it removes their high-frequency components and denoises them.
The output of $\text{DWT}_{ll}$, i.e., the low-frequency component, saves the main information of the feature map
to extract the identifiable features.
During down-sampling of WaveCNets, $\text{DWT}_{ll}$ could resist the noise propagation in the deep networks
and helps to maintain the basic object structure in the feature maps.
Therefore, $\text{DWT}_{ll}$ would accelerate the training of deep networks and
lead to better noise-robustness and increased classification accuracy.

\section{Experiments}
\begin{table*}[!t]
	\scriptsize
	\caption{Top-1 accuracy of WaveCNets on ImageNet validation set.}
	\label{Tab_WaveCNet_accuracy}
	\begin{center}
	\begin{threeparttable}
		\setlength{\tabcolsep}{1.75mm}{
			\begin{tabular}{cc||c|cccc|c}\hline
		\multicolumn{2}{c||}{Wavelet}&{WVGG16bn} &WResNet18&WResNet34&WResNet50&WResNet101 &{WDenseNet121} \\\hline\hline
\multicolumn{2}{c||}{None (baseline)\tnote{*}}
							&73.37		     &69.76		&73.30		&76.15		&77.37		&74.65		\\\hline
\multicolumn{2}{c||}{Haar}
	 						&74.10 ($+$0.73) &71.47 ($+$1.71)	&74.35 ($+$1.05)	&\textbf{76.89 ($+$0.74)}&78.23 ($+$0.86)	&75.27 ($+$0.62)	\\\hline
{\multirow{4}{*}{Cohen}}
&\multicolumn{1}{|c||}{ch2.2}&74.31 ($+$0.94)&\textbf{71.62 ($+$1.86)}	&74.33 ($+$1.03)	&76.41 ($+$0.26)&78.34 ($+$0.97)	&75.36 ($+$0.71)	\\
&\multicolumn{1}{|c||}{ch3.3}&\textbf{74.40 ($+$1.03)}&71.55 ($+$1.79)	&74.51 ($+$1.21)	&76.71 ($+$0.56)&\textbf{78.51 ($+$1.14)}	&\textbf{75.44 ($+$0.79)}\\
&\multicolumn{1}{|c||}{ch4.4}&74.02 ($+$0.65)&71.52 ($+$1.76)	&\textbf{74.61 ($+$1.31)}	&76.56 ($+$0.41)&78.47 ($+$1.10)	&75.29 ($+$0.64)	\\
&\multicolumn{1}{|c||}{ch5.5}&73.67 ($+$0.30)&71.26 ($+$1.50)	&74.34 ($+$1.04)	&76.51 ($+$0.36)	&78.39 ($+$1.02)	&75.01 ($+$0.36)	\\\hline
{\multirow{5}{*}{Daubechies}}&
\multicolumn{1}{|c||}{db2} 	 &74.08 ($+$0.71)&71.48 ($+$1.72)	&74.30 ($+$1.00)	&76.27 ($+$0.12)	&78.29 ($+$0.92)	&75.08 ($+$0.43)	\\
&\multicolumn{1}{|c||}{db3}	 &				 &71.08 ($+$1.32)	&74.11 ($+$0.81)	&76.38 ($+$0.23)	&					&					\\
&\multicolumn{1}{|c||}{db4}	 &				 &70.35 ($+$0.59)	&73.53 ($+$0.23)	&75.65 ($-$0.50)	&					&					\\
&\multicolumn{1}{|c||}{db5}	 &				 &69.54 ($-$0.22)	&73.41 ($+$0.11)	&74.90 ($-$1.25)	&					&					\\
&\multicolumn{1}{|c||}{db6}	 &				 &68.74 ($-$1.02)	&72.68 ($-$0.62)	&73.95 ($-$2.20)	&					&					\\\hline
			\end{tabular}}
		\begin{tablenotes}
			\item[*] corresponding to the results of original CNNs, i.e., VGG16bn, ResNets, DenseNet121.
		\end{tablenotes}
	\end{threeparttable}
	\end{center}
\end{table*}
The commonly used CNN architectures for image classification,
such as VGG \cite{simonyan2014very}, ResNets \cite{he2016deep}, DenseNet \cite{huang2017densely},
compose of various max-pooling, average-pooling, and strided-convolution.
By upgrading the down-sampling with Eqs. (\ref{eq_maxpool_up}) - (\ref{eq_avgpool_up}),
we create WaveCNets, including WVGG16bn, WResNets, WDenseNet121.
Compared with the original CNNs, WaveCNets do not employ additional learnable parameters.
We evaluate their classification accuracies and noise-robustness
using ImageNet \cite{deng2009imagenet} and ImageNet-C \cite{hendrycks2019benchmarking}.
At last, we explore the potential of wavelet integrated deep networks for image segmentation.

\subsection{ImageNet classification}
ImageNet contains 1.2M training and 50K validation images from 1000 categories.
On the training set,
we train WaveCNets when various wavelets are used,
with the standard training protocols from the publicly available PyTorch \cite{paszke2017automatic} repository.
Table \ref{Tab_WaveCNet_accuracy} presents the top-1 accuracy of WaveCNets on ImageNet validation set,
where ``haar'', ``dbx'', and ``chx.y'' denote the Haar wavelet, Daubechies wavelet with approximation order $x$,
and Cohen wavelet with orders $(x,y)$.
The length of the wavelet filters increase as the orders increase.
While Haar and Cohen wavelets are symmetric, Daubechies are not.

In Table \ref{Tab_WaveCNet_accuracy}, parenthesized numbers are accuracy difference compared with the baseline results.
The baseline results, i.e., the results of the original CNNs, are sourced from the official PyTorch \cite{paszke2017automatic}.
For all CNN architectures, Haar and Cohen wavelets improve their classification accuracy,
although the best wavelet varies with CNN.
For example, Haar and Cohen wavelets improve the accuracy of ResNet18 by $1.50\%$ to $1.86\%$.
However, the performance of asymmetric Daubechies wavelet gets worse as the approximation order increases.
Daubechies wavelets with shorter filters (``db2'' and ``db3'') could improve the CNN accuracy,
while that with longer filters (``db5'' and ``db6'') may reduce the accuracy.
For example, the top-1 accuracy of WResNet18 decreases from $71.48\%$ to $68.74\%$.
We conclude that the symmetric wavelets perform better than asymmetric ones in image classification.
That is the reason why we do not train WVGG16bn, WResNet101, WDenseNet121 with ``db3'', ``db4'', ``db5'', ``db6''.

\begin{figure}[!bpt]
	\centering
	\includegraphics*[scale=0.55, viewport=21 2 417 305]{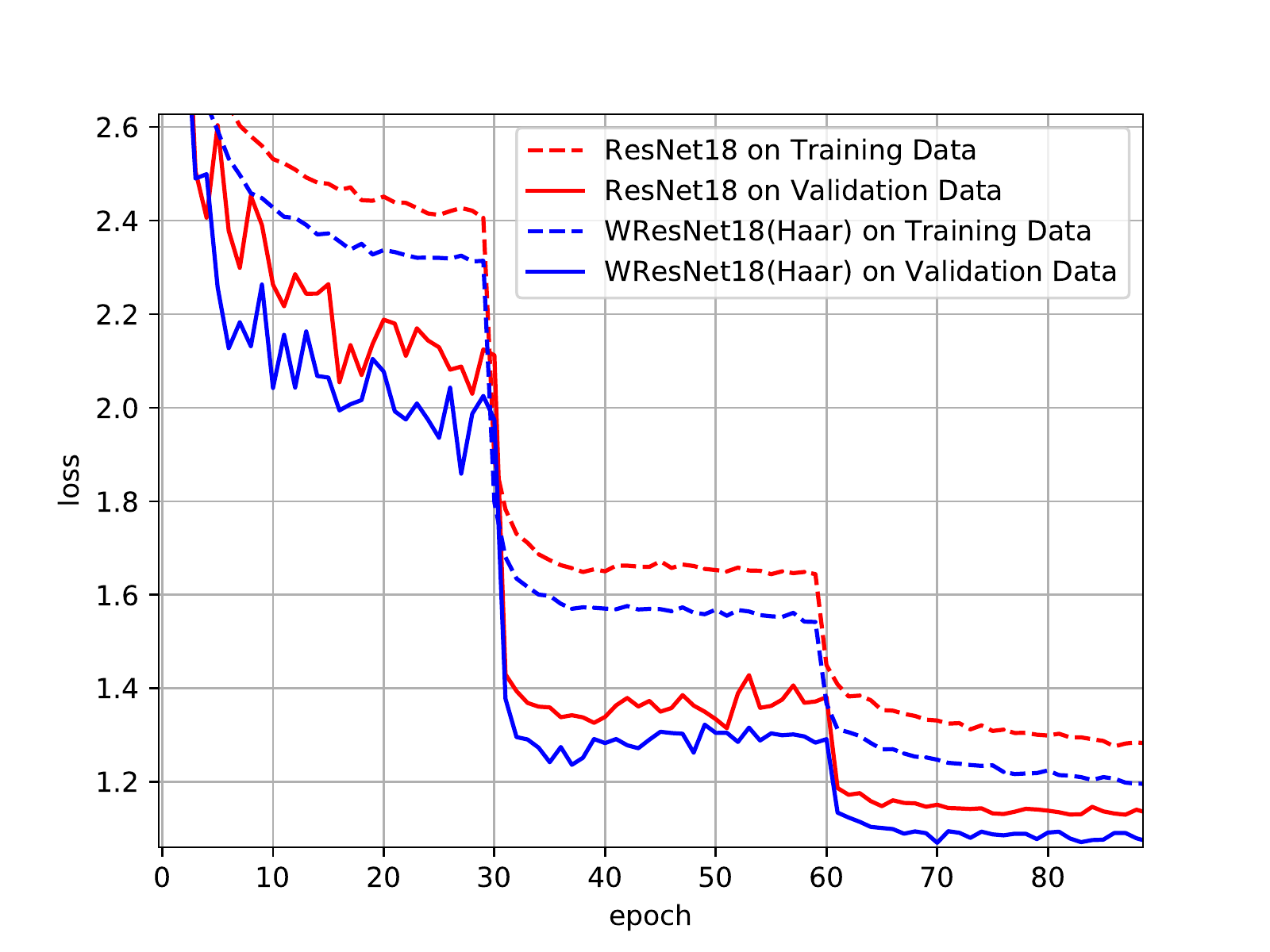}
	\caption{The loss of ResNet18 and WResNet18(Haar).}
	\label{fig_loss_resnet18}
\end{figure}
\begin{figure*}[!bpt]
	\centering
	\subfigure[VGG16bn and WVGG16bn]
	{\includegraphics*[scale=0.22, viewport=175 65 1070 604]{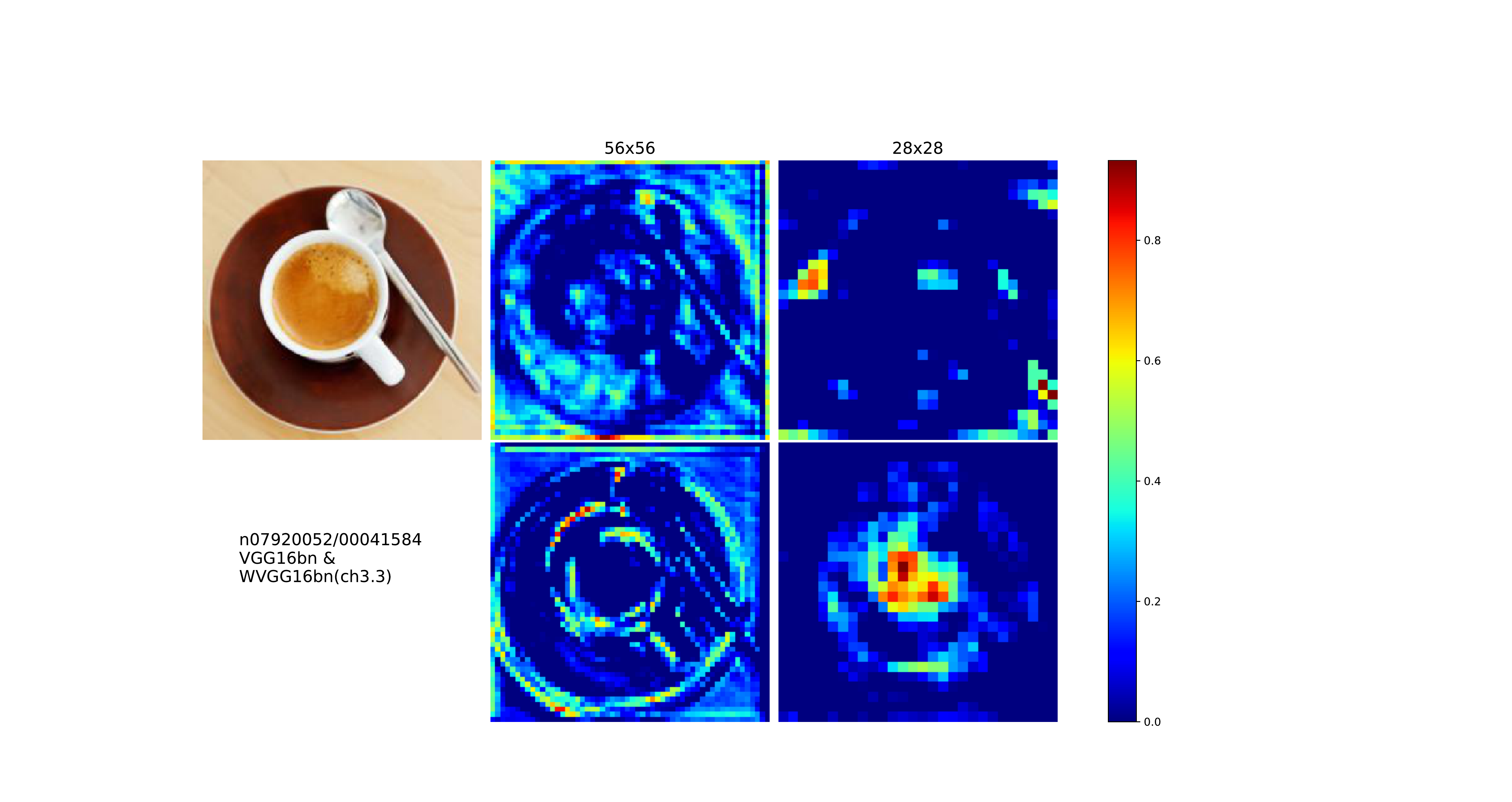}\label{fig_feature_maps_a}}\hspace{30pt}
	\subfigure[ResNet18 and WResNet18]
	{\includegraphics*[scale=0.22, viewport=175 65 1070 604]{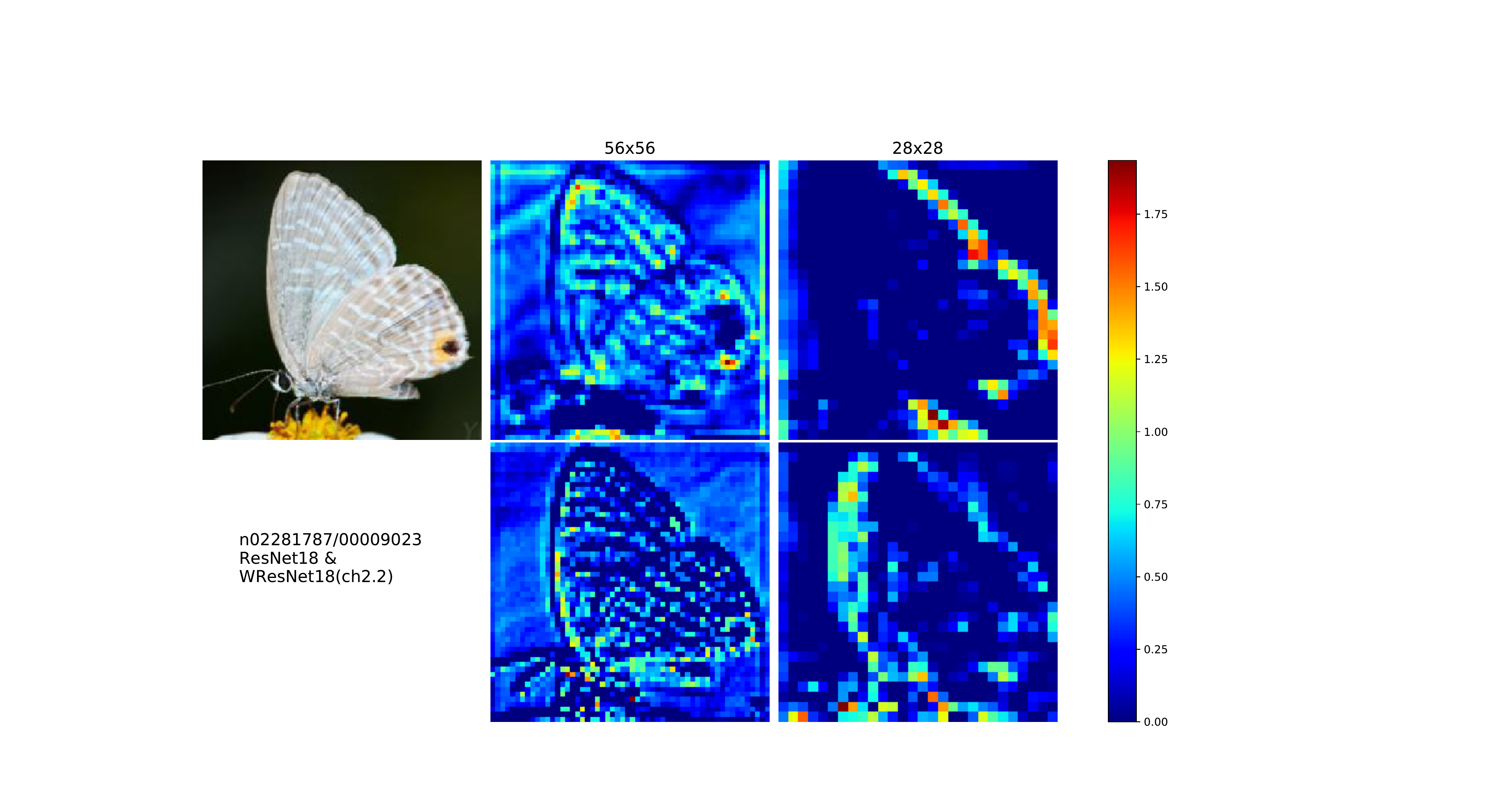}\label{fig_feature_maps_b}}\\
	\subfigure[ResNet34 and WResNet34]
	{\includegraphics*[scale=0.22, viewport=175 65 1070 604]{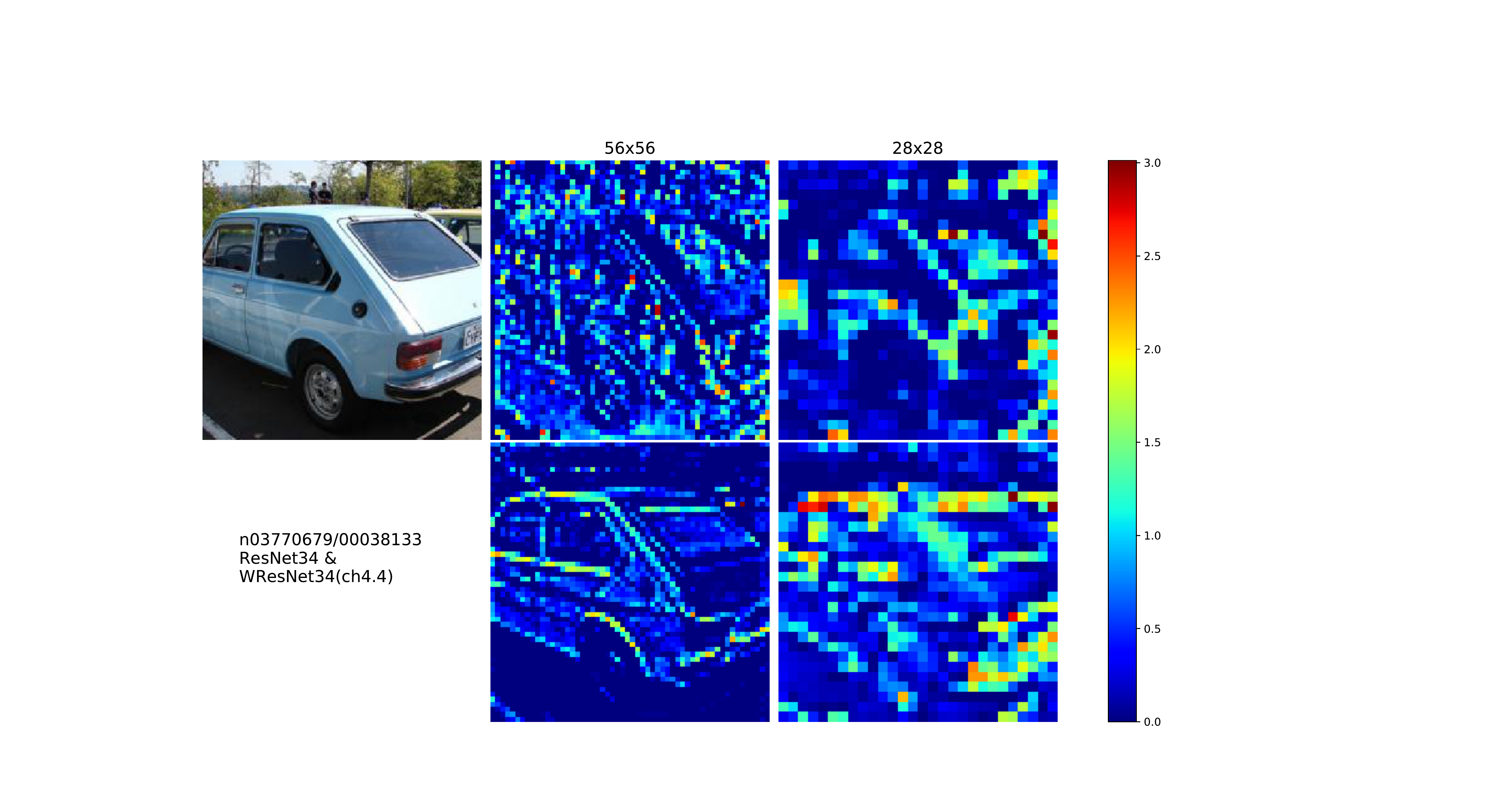}\label{fig_feature_maps_c}}\hspace{30pt}
	\subfigure[ResNet50 and WResNet50]
	{\includegraphics*[scale=0.22, viewport=175 65 1070 604]{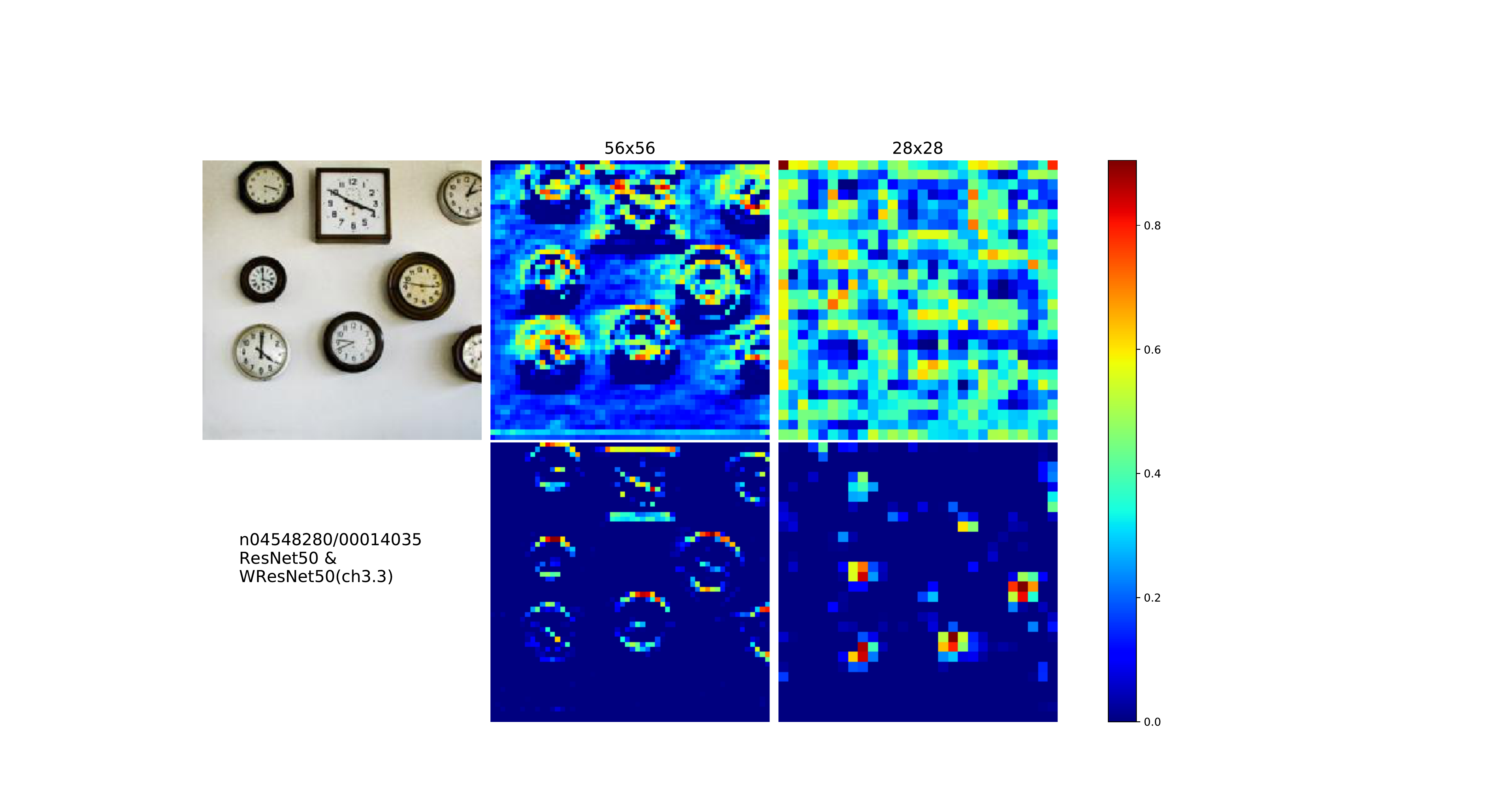}\label{fig_feature_maps_d}}
	\caption{The feature maps of CNNs (top) and WaveCNets (bottom).}
	\label{fig_feature_maps}
\end{figure*}
We retrain ResNet18 using the standard ImageNet classification training repository in PyTorch.
In Fig. \ref{fig_loss_resnet18}, we compare the losses of ResNet18 and WResNet18(Haar) during the training procedure.
Fig. \ref{fig_loss_resnet18} adopts red dashed and green dashed lines to denote the train losses of ResNet18 and WResNet18(Haar), respectively.
Throughout the whole training procedure, the training loss of WResNet18(Haar) is about $0.08$ lower than that of ResNet18,
when the two networks employ the same amount of learnable parameters.
This suggests that wavelet accelerates the training of ResNet18 architecture.
On the validation set, WResNet18 loss (green solid line) is also always lower than ResNet18 loss (red solid line),
which lead to the increase of final classification accuracy by $1.71\%$.

Fig. \ref{fig_feature_maps} presents four example feature maps of well trained CNNs and WaveCNets.
In each subfigure, the top row shows the input image with size of $224\times224$ from ImageNet validation set
and the two feature maps produced by original CNN,
while the bottom row shows the related information (image, CNN and WaveCNet names) and feature maps produced by the WaveCNet.
The two feature maps are captured from the 16th output channel of the final layer
in the network blocks with tensor size of $56\times56$ (middle) and $28\times28$ (right), respectively.
The feature maps have been enlarged for better illustration.

From Fig. \ref{fig_feature_maps}, one can find
that the backgrounds of the feature map produced by WaveCNets are cleaner than that produced by CNNs,
and the object structures in the former are more complete than that in the latter.
For example, in the top row of Fig. \ref{fig_feature_maps_d},
the clock boundary in the ResNet50 feature map with size of $56\times56$ are fuzzy,
and the basic structures of clocks have been totally broken by strong noise in the feature map with size of $28\times28$.
In the second row,
the backgrounds of feature maps produced by WResNet50(ch3.3) are very clean,
and it is easy to figure out the clock structures in the feature map with size of $56\times56$
and the clock areas in the feature map with size of $28\times28$.
The above observations illustrate that the down-sampling operations could cause noise accumulation
and break the basic object structures during CNN inference,
while DWT in WaveCNets relieves these drawbacks.
We believe that this is the reason why WaveCNets converge faster in training and ultimately achieve better classification accuracy.

In \cite{zhang2019making}, the author is surprised at the increased classification accuracy of CNNs
after filtering is integrated into the down-sampling.
In \cite{geirhos2018imagenet},
the authors show that ``ImageNet-trained CNNs are strongly biased towards recognising textures rather than shapes''.
Our experimental results suggest that this may be sourced from the commonly used down-sampling operations,
which tend to break the object structures and accumulate noise in the feature maps.

\begin{figure}[bpt]
	\centering
	\includegraphics*[scale=0.5, viewport=23 8 418 370]{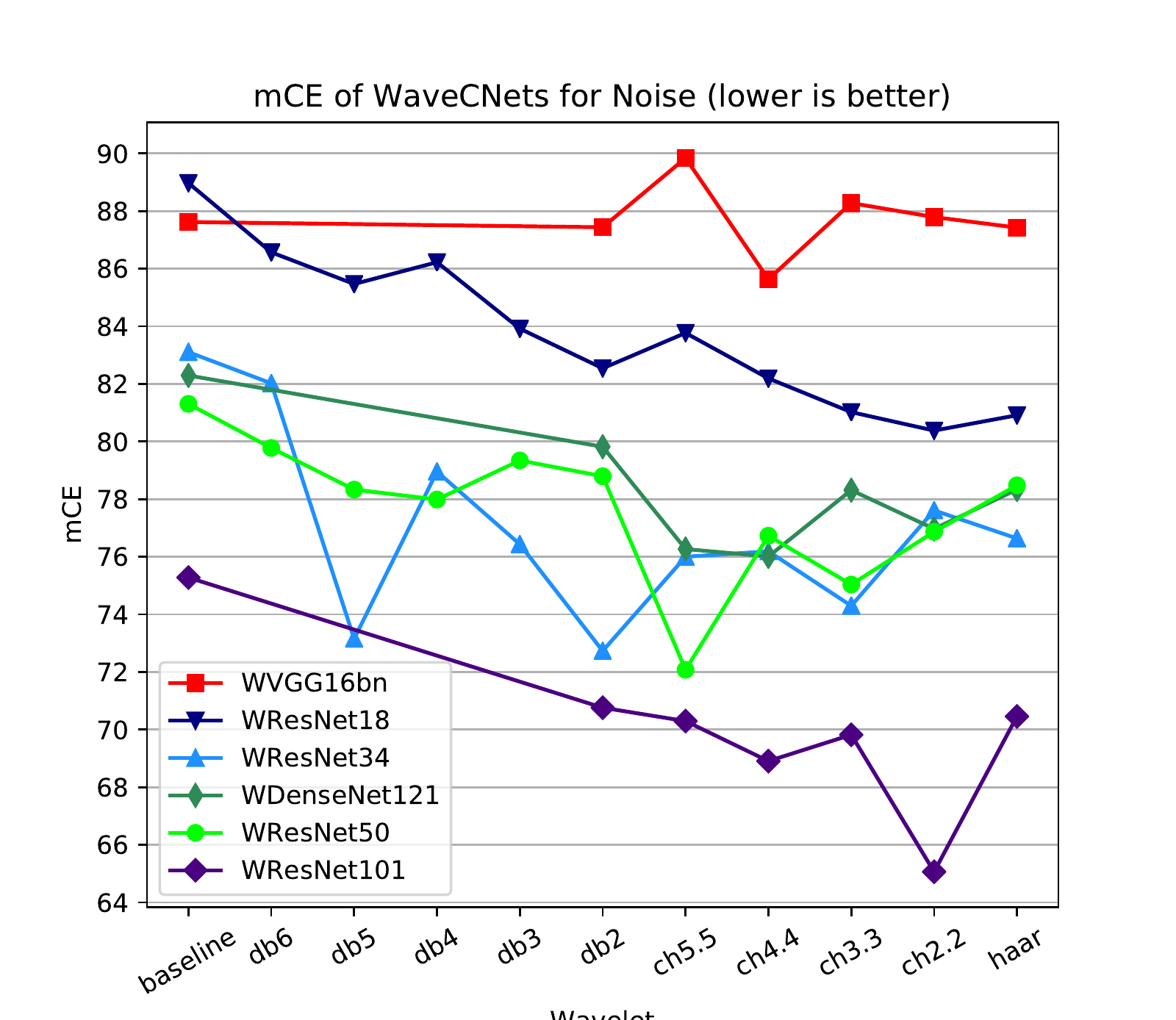}
	\caption{The noise mCE of WaveCNets.}
	\label{fig_mCE_noise}
\end{figure}
\subsection{Noise-robustness}
In \cite{hendrycks2019benchmarking},
the authors corrupt the ImageNet validation set using 15 visual corruptions with five severity levels,
to create ImageNet-C and test the robustness of ImageNet-trained classifiers to the input corruptions.
The 15 corruptions are sourced from four categories,
i.e., noise (Gaussian noise, shot noise, impulse noise),
blur (defocus blur, frosted glass blur, motion blur, zoom blur),
weather (snow, frost, fog, brightness), and digital (contrast, elastic, pixelate, JPEG-compression).
$E_{s,c}^f$ denotes the top-1 error of a trained classifier $f$ on corruption type $c$ at severity level $s$.
The authors present the Corruption Error $\text{CE}_c^f$, computed with
\begin{align}
\label{eq_CE}
\text{CE}_c^f &= \sum_{s=1}^5 E_{s,c}^f\left/\sum_{s=1}^5 E_{s,c}^{\text{AlexNet}}\right.,
\end{align}
to evaluate the performance of a trained classifier $f$.
In Eq. (\ref{eq_CE}), the authors normalize the error using the top-1 error of AlexNet \cite{krizhevsky2012imagenet}
to adjust the difference of various corruptions.

In this section, we use the noise part (750K images, 50K $\times$ 3 $\times$ 5) of ImageNet-C
and
\begin{align}
\label{eq_mCE_noise}
\text{mCE}_{\text{noise}}^f &= \dfrac{1}{3}\left(\text{CE}_{\text{Gaussian}}^f
+ \text{CE}_{\text{shot}}^f + \text{CE}_{\text{impulse}}^f\right)
\end{align}
to evaluate the noise-robustness of WaveCNet $f$.

\begin{figure*}[bpt]
	\centering
	\subfigure[Gaussian noise]
	{\includegraphics*[scale=0.22, viewport=175 65 1062 608]{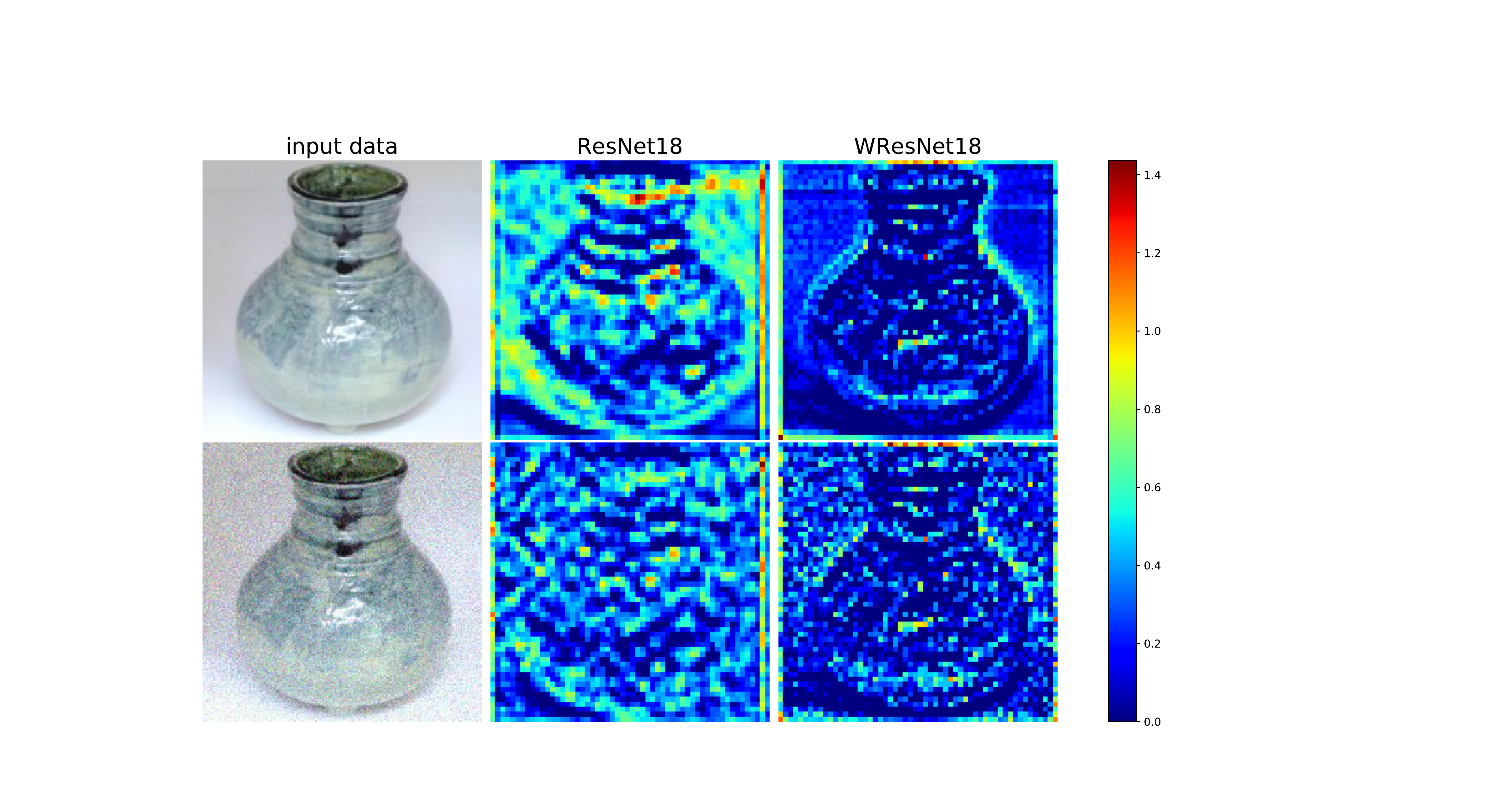}
	\label{fig_feature_map_noise_a}}\hspace{30pt}
	\subfigure[Impulse noise]
	{\includegraphics*[scale=0.22, viewport=175 65 1068 608]{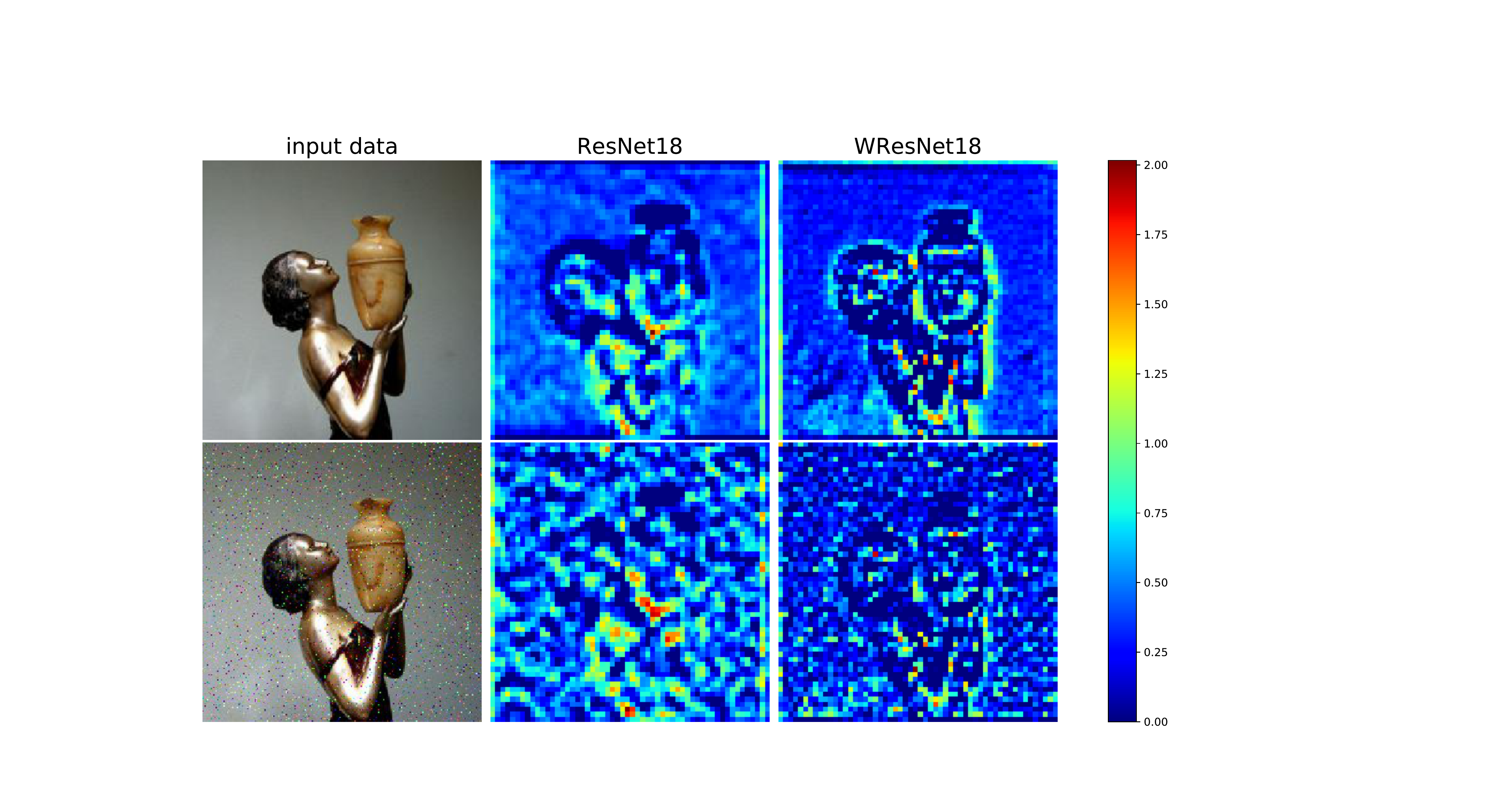}
	\label{fig_feature_map_noise_b}}
	\caption{The feature maps sourced from clean (top) and noisy (bottom) images.}
	\label{fig_feature_map_noise}
\end{figure*}
We test the top-1 errors of WaveCNets and AlexNet on each noise corruption type $c$ at each level of severity $s$,
when WaveCNets and AlexNet are trained on the clean ImageNet training set.
Then, we compute $\text{mCE}_{\text{noise}}^{\text{WaveCNet}}$ according to Eqs. (\ref{eq_CE}) and (\ref{eq_mCE_noise}).
In Fig. \ref{fig_mCE_noise}, we show the noise mCEs of WaveCNets for different network architectures and various wavelets.
The ``baseline'' corresponds to the noise mCEs of original CNN architectures,
while ``dbx'', ``chx.y'' and ``haar'' correspond to the mCEs of WaveCNets with different wavelets.
Except VGG16bn, our method obviously increase the noise-robustness of the CNN architectures for image classification.
For example, the noise mCE of ResNet18 (with navy blue color and down triangle marker in Fig. \ref{fig_mCE_noise})
decreases from $88.97$ (``baseline'') to $80.38$ (``ch2.2'').
One can find that the all wavelets including ``db5'' and ``db6'' improve the noise-robustness of ResNet18, ResNet34, and ResNet50,
although the classification accuracy of the WResNets with ``db5'' and ``db6'' for the clean images may be lower than that of the original ResNets.
It means that our methods indeed increase the noise-robustness of these network architectures.

Fig. \ref{fig_feature_map_noise} shows two example feature maps for well trained ResNet18 and WResNet18
with noisy images as input.
In every subfigure, the first row shows the clean image with size of $224\times224$ from ImageNet validation set
and feature maps generated by ResNet18 and WResNet18(ch2.2), respectively.
The second row shows the image added with Gaussian or impulse noise and the feature maps generated by the two networks.
These feature maps are captured from the 16th output channel of the last layer in the network blocks with tensor size of $56\times56$.
From the two examples, one can find that it is difficult for the original CNN to suppress noise,
while WaveCNet could suppress the noise and maintain the object structure during its inference.
For example, in Fig. \ref{fig_feature_map_noise_a},
the bottle structures in the two feature maps generated by ResNet18 and WResNet18(ch2.2) are complete,
when the clean porcelain bottle image is fed into the networks.
However, after the image is corrupted by Gaussian noise,
the ResNet18 feature map contains very strong noise and the bottle structure vanishs,
while the basic structure could still be observed from the WResNet18 feature map.
This advantage improves the robustness of WaveCNets against different noise.

The noise-robustness of VGG16bn is inferior to that of ResNet34,
although they achieve similar accuracy ($73.37\%$ and $73.30\%$).
Our method can not significantly improve the noise-robustness of VGG16bn,
although it can increase the accuracy by $1.03\%$.
It means that the VGG16bn may be not a proper architecture in terms of noise-robustness.

\subsection{Comparison with other wavelet based down-sampling}
\begin{figure}[bpt]
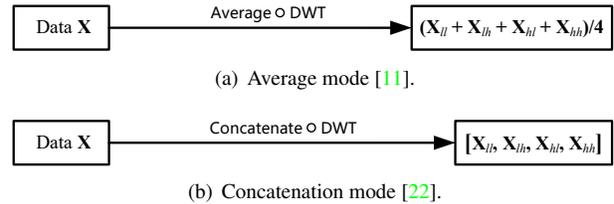

	\centering
	\subfigure[Average mode \cite{duan2017sar}.]{
		\label{fig_wavelet_integrated_modes_b}
		\includegraphics*[scale=0.725, viewport=267 130 585 155]{figures/Visio-down_sampling_of_dwt.pdf}
	}\\
	\subfigure[Concatenation mode \cite{liu2018multi}.]{
		\label{fig_wavelet_integrated_modes_a}
		\includegraphics*[scale=0.725, viewport=267 179 585 204]{figures/Visio-down_sampling_of_dwt.pdf}
	}
	\caption{Wavelet integrated down-sampling in various modes.}
	\label{fig_wavelet_integrated_modes}
\end{figure}
Different with our DWT based down-sampling (Fig. \ref{fig_denoise_b}),
there are other wavelet integrated down-sampling modes in literatures.
In \cite{duan2017sar}, the authors adopt as down-sampling output the average value of the multiple components of wavelet transform,
as Fig. \ref{fig_wavelet_integrated_modes_b} shows.
In \cite{liu2018multi}, the authors concatenate all the components output from DWT, and process them in a unified way,
as Fig. \ref{fig_wavelet_integrated_modes_a} shows.

Here, taking ResNet18 as backbone, we compare our wavelet integrated down-sampling with the previous approaches,
in terms of classification accuracy and noise-robustness.
We rebuild ResNet18 using the three down-sampling modes shown in Fig. \ref{fig_denoise_b} and Fig. \ref{fig_wavelet_integrated_modes},
and denote them as WResNet18, WResNet18\_A, and WResNet18\_C, respectively.
We train them on ImageNet when various wavelets are used.
Table \ref{Tab_result_other_modes} shows the accuracy on ImageNet
and the noise mCEs on the ImageNet-C.
Generally, the networks using wavelet based down-sampling achieve better accuracy and noise mCE
than that of original network, ResNet18 ($69.76\%$ accuracy and $88.97$ mCE).
\begin{table}
\scriptsize
	\caption{Comparison with other wavelet based down-sampling.}
	\label{Tab_result_other_modes}
\begin{center}
\setlength{\tabcolsep}{1.00mm}{
\begin{tabular}{l|cccccc|c}\hline
\multicolumn{1}{c|}{\multirow{2}{*}{Network}} &\multicolumn{6}{c|}{Top-1 Accuracy (higher is better)} & \multirow{2}{*}{Params.}\\\cline{2-7}
						 					&haar	&ch2.2	&ch3.3	&ch4.4	&ch5.5	&db2 	&\\\hline
					     WResNet18			&71.47	&71.62		&71.55		&71.52		&71.26		&71.48	&\textbf{11.69}M\\
		WResNet18\_A \cite{duan2017sar} 	&70.06	&69.24		&69.91		&69.98		&70.31		&70.52	&11.69M\\
		WResNet18\_C \cite{liu2018multi} 	&\textbf{71.94}	&\textbf{71.75}		&\textbf{71.66}		&\textbf{71.99}		&\textbf{72.03}		
						 					&\textbf{71.88}	&21.62M\\\cdashline{1-8}[2pt/2pt]
						 WResNet34			&74.35	&74.33		&74.51		&74.61		&74.34		&74.30	&21.80M\\\hline\hline
						 \multirow{2}{*}{ } &\multicolumn{6}{c|}{Noise mCE (lower is better)} & \\\hline
						 WResNet18 			&\textbf{80.91}	&\textbf{80.38}		&\textbf{81.02}		&82.19		&83.77		&82.54	& \\
		WResNet18\_A \cite{duan2017sar} 	&83.17	&86.02		&86.07		&85.22		&82.96		&84.01	& \\
		WResNet18\_C \cite{liu2018multi} 	&81.79	&83.67		&83.51		&\textbf{82.13}		&\textbf{82.60}		
						 					&\textbf{80.11}	& \\\cdashline{1-8}[2pt/2pt]
						 WResNet34			&76.64	&77.61		&74.30		&76.19		&76.00		&72.73	& \\\hline
\end{tabular}}
\end{center}
\end{table}

Similar to WResNet18, the number of parameters of WResNet18\_A is the same with that of original ResNet18.
However, the added high-frequency components in the feature maps damage the information contained in the low-frequency component,
because of the high-frequency noise.
WResNet18\_A performs the worst among the networks using wavelet based down-sampling.

Due to the tensor concatenation,
WResNet18\_C employs much more parameters ($21.62 \times 10^6$) than WResNet18 and WResNet18\_A ($11.69 \times 10^6$).
WResNet18\_C thus increase the accuracy of WResNet18 by $0.11\%$ to $0.77\%$, when various wavelets are used.
However, due to the included noise,
the concatenation does not evidently improve the noise-robustness.
In addition, the amount of parameters for WResNet18\_C is almost the same with that for WResNet34 ($21.80\times10^6$),
while the accuracy and noise mCE of WResNet34 are obviously superior to that of WResNet18\_C.


\subsection{Image segmentation}
The main contributions of our method are the DWT and IDWT layers.
IDWT is a useful up-sampling approach to recover the data details.
With IDWT, WaveCNets can be easily transferred to image-to-image translation tasks.
We now test their applications in semantic image segmentation.

\begin{figure}[bpt]
	\centering
	\subfigure[SegNet]{
		\label{fig_dual_structure_Pooling_Unpooling}
		\includegraphics*[scale=0.5, viewport=72 479 285 561]{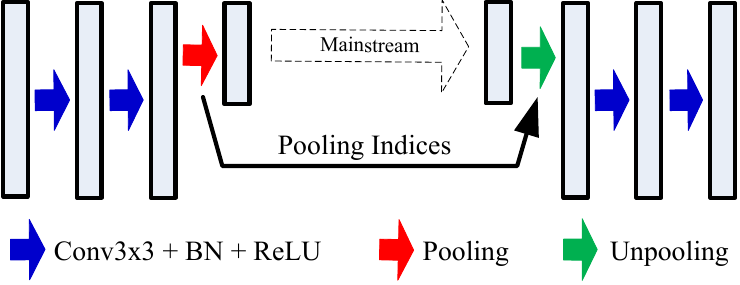}
	}\hspace{10pt}
	\subfigure[WaveUNets]{
		\label{fig_dual_structure_DWT_IDWT}
		\includegraphics*[scale=0.5, viewport=72 380 285 462]{figures/dual_structures.pdf}
	}\hspace{25pt}
	\caption{Down-sampling and up-sampling used in SegNet and WaveUNet.}\label{fig_dual_structures}
\end{figure}
To restore details in image segmentation, we design WaveUNets
by replacing the max-pooling and max-unpooling in SegNet \cite{badrinarayanan2017segnet} with DWT and IDWT.
SegNet adopts encoder-decoder architecture and uses VGG16bn as its encoder backbone.
In its decoder, SegNet recovers the feature map resolution using max-unpooling,
as Fig. \ref{fig_dual_structure_Pooling_Unpooling} shows.
While max-unpooling only recover very limited details,
IDWT can recover most of the data details.
In the encoder, WaveUNets decompose the feature maps into various frequency components,
as Fig. \ref{fig_dual_structure_DWT_IDWT} shows.
While the low-frequency components are used to extract high-level features,
the high-frequency components are stored and transmitted to the decoder for
resolution restoration with IDWT.

We evaluate WaveUNets and SegNet using CamVid \cite{brostow2009semantic} dataset.
CamVid contains 701 road scene images (367, 101, and 233 for the training, validation, and test).
In \cite{badrinarayanan2017segnet}, the authors train the SegNet using an extended CamVid training set containing 3433 images,
which achieved $60.10\%$ mIoU on the CamVid test set.
We train SegNet and WaveUNet with various wavelets using only 367 CamVid training images.
Table \ref{Tab_IoU_CamVid} shows the mIoU on the CamVid test set.
Our WaveUNets get higher mIoU and achieve the best result ($64.23\%$) with Haar wavelet.

\begin{figure}[bpt]
	\centering
	\includegraphics*[scale=0.7, viewport=40 616 382 776]{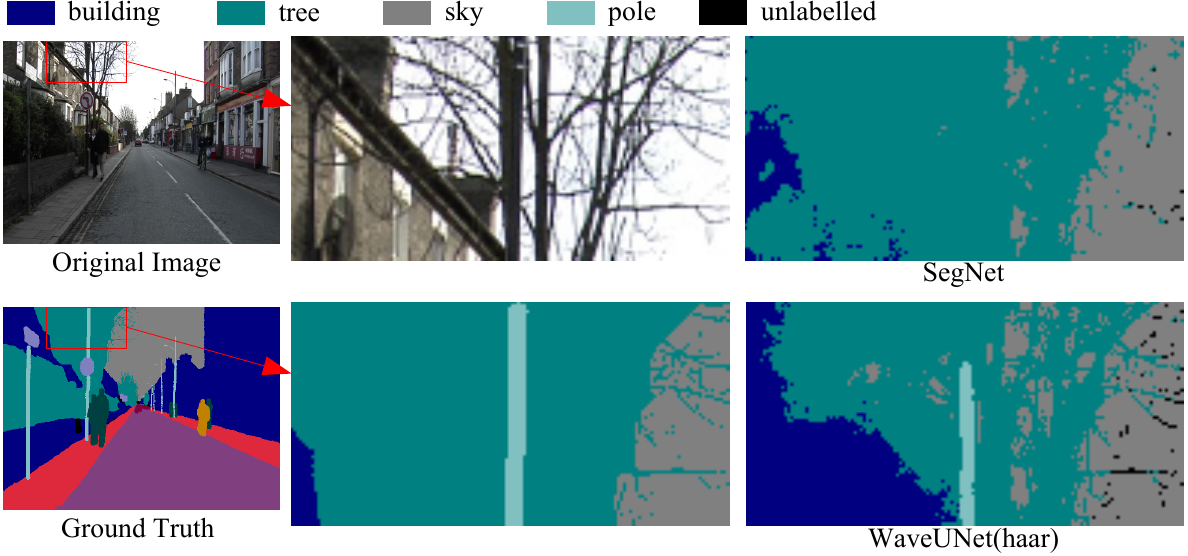}
	\caption{Comparison of SegNet and WaveUNet segmentations.}
	\label{Fig_Segmentation_Comparison}
\end{figure}

In Fig. \ref{Fig_Segmentation_Comparison}, we present a visual example for SegNet and WaveUNet segmentations.
Fig. \ref{Fig_Segmentation_Comparison} shows the example image, its manual annotation,
a region consisting of ``building'', ``tree'', ``sky'' and ``pole'', and the segmentation results achieved using SegNet and WaveUNet.
The region has been enlarged with colored segmentation results for better illustration.
From the figure, one can find in the segmentation result that WaveUNet keeps the basic structure of ``tree'', ``pole'', and ``building''
and restores the object details, such as the ``tree'' branches and the ``pole''.
The segmentation result of WaveUNet matches the image region much better than that of SegNet,
even corrects the annotation noise about ``building'' and ``tree'' in the ground truth.
\begin{table}[!t]
	\scriptsize
	\caption{Results on CamVid test set.}
	\label{Tab_IoU_CamVid}
	\begin{center}
	\begin{threeparttable}
	\setlength{\tabcolsep}{1.00mm}{
	\begin{tabular}{r|cc|cccccc}\hline
	\multirow{2}{*}{Network} &  \multicolumn{2}{c|}{SegNet} &  \multicolumn{6}{c}{Our WaveUNets} \\\cline{2-9}
    & {\cite{badrinarayanan2017segnet}} &  Ours &haar&ch2.2&ch3.3&ch4.4&ch5.5&db2       \\ \hline
	mIoU & 60.10  & 57.89& \textbf{64.23}  &   63.35 &  62.90 &  63.76 & 63.61 &  63.78\\ \hline
	\end{tabular}}
	\end{threeparttable}
	\end{center}
\end{table}


\section{Conclusions}
We transform Discrete Wavelet Transform (DWT) and Inverse DWT (IDWT) into general network layers,
and design wavelet integrated convolutional networks (WaveCNets) for image classification.
Being able to well keep object structures and suppress data noise during network inference,
WaveCNets achieve higher image classification accuracy and better noise-robustness
for various commonly used network architectures.


\section*{Acknowledgments}
The work was supported by the Natural Science Foundation of China under grants no. 61672357, 91959108 and U1713214,
and the Science and Technology Project of Guangdong Province under grant no. 2018A050501014.

{\small
\bibliographystyle{ieee_fullname}
\bibliography{egbib}
}

\clearpage
\section*{Supplementary materials for ``Wavelet Integrated CNNs for Noise-Robust Image Classification''}\label{APP_wavelet}
\subsection*{A. Wavelets}
\begin{table*}
	\scriptsize
	\caption{Low-pass filters of the Daubechies wavelets.}
	\label{Tab_Daubechies_banks}
	\begin{center}
	\setlength{\tabcolsep}{1mm}{
	\begin{tabular}{c|cccccc}
		\hline
		          $p$           &     $1$      &       $2$       &         $3$         &         $4$         &         $5$         &         $6$         \\ \hline
		\multirow{12}{*}{$l_k$} &     $1$      &  $1+\sqrt{3}$   & $~~~0.332670552950$ & $~~~0.230377813309$ & $~~~0.160102397974$ & $~~~0.111540743350$ \\
		                        &     $1$      &  $3+\sqrt{3}$   & $~~~0.806891509311$ & $~~~0.714846570553$ & $~~~0.603829269797$ & $~~~0.494623890398$ \\
		                        &              &  $3-\sqrt{3}$   & $~~~0.459877502118$ & $~~~0.630880767930$ & $~~~0.724308528438$ & $~~~0.751133908021$ \\
		                        &              &  $1-\sqrt{3}$   &  $-0.135011020010$  &  $-0.027983769417$  & $~~~0.138428145901$ & $~~~0.315250351709$ \\
		                        &              &                 &  $-0.085441273882$  &  $-0.187034811719$  &  $-0.242294887066$  &  $-0.226264693965$  \\
		                        &              &                 & $~~~0.035226291886$ & $~~~0.030841381836$ &  $-0.032244869585$  &  $-0.129766867567$  \\
		                        &              &                 &                     & $~~~0.032883011667$ & $~~~0.077571493840$ & $~~~0.097501605587$ \\
		                        &              &                 &                     &  $-0.010597401785$  &  $-0.006241490213$  & $~~~0.027522865530$ \\
		                        &              &                 &                     &                     &  $-0.012580751999$  &  $-0.031582039317$  \\
		                        &              &                 &                     &                     & $~~~0.003335725285$ & $~~~0.000553842201$ \\
		                        &              &                 &                     &                     &                     & $~~~0.004777257511$ \\
		                        &              &                 &                     &                     &                     &  $-0.001077301085$  \\ \hline
		       $\text{factor}$         & $1/\sqrt{2}$ & $1/(4\sqrt{2})$ &         $1$         &         $1$         &         $1$         &         $1$         \\ \hline
	\end{tabular}}
	\end{center}
\end{table*}
\begin{table*}
\scriptsize
\caption{Low-pass filters of the Cohen wavelets.}
\label{Tab_CDF_banks}
\begin{center}
	\setlength{\tabcolsep}{0.5mm}{
		\begin{tabular}{c|cc|cc|cc|cc}
			\hline
			$(p,\tilde{p})$     & \multicolumn{2}{|c|}{$(2,2)$}       & \multicolumn{2}{|c|}{$(3,3)$}       & \multicolumn{2}{|c|}{$(4,4)$}          & \multicolumn{2}{|c}{$(5,5)$}           \\ \hline
			banks          & $\textbf{l}$ & $\tilde{\textbf{l}}$ & $\textbf{l}$ & $\tilde{\textbf{l}}$ &  $\textbf{l}$   & $\tilde{\textbf{l}}$ &  $\textbf{l}$   & $\tilde{\textbf{l}}$ \\ \hline
			\multirow{12}{*}{$l_k$} &     $0$      &         $0$          &     $0$      &   $~~~0.06629126$    &       $0$       &         $0$          & $~~~0.01345671$ &         $0$          \\
			& $0.35355339$ &    $-0.17677670$     &     $0$      &    $-0.19887378$     &  $-0.06453888$  &   $~~~0.03782846$    &  $-0.00269497$  &         $0$          \\
			& $0.70710678$ &   $~~~0.35355339$    & $0.17677670$ &    $-0.15467961$     &  $-0.04068942$  &    $-0.02384947$     &  $-0.13670658$  &   $~~~0.03968709$    \\
			& $0.35355339$ &   $~~~1.06066017$    & $0.53033009$ &   $~~~0.99436891$    & $~~~0.41809227$ &    $-0.11062440$     &  $-0.09350470$  &   $~~~0.00794811$    \\
			&     $0$      &   $~~~0.35355339$    & $0.53033009$ &   $~~~0.99436891$    & $~~~0.78848562$ &   $~~~0.37740286$    & $~~~0.47680327$ &    $-0.05446379$     \\
			&     $0$      &    $-0.17677670$     & $0.17677670$ &    $-0.15467961$     & $~~~0.41809227$ &   $~~~0.85269868$    & $~~~0.89950611$ &   $~~~0.34560528$    \\
			&              &                      &     $0$      &    $-0.19887378$     &  $-0.04068942$  &   $~~~0.37740286$    & $~~~0.47680327$ &   $~~~0.73666018$    \\
			&              &                      &     $0$      &   $~~~0.06629126$    &  $-0.06453888$  &    $-0.11062440$     &  $-0.09350470$  &   $~~~0.34560528$    \\
			&              &                      &              &                      &       $0$       &    $-0.02384947$     &  $-0.13670658$  &    $-0.05446379$     \\
			&              &                      &              &                      &       $0$       &   $~~~0.03782846$    &  $-0.00269497$  &   $~~~0.00794811$    \\
			&              &                      &              &                      &                 &                      & $~~~0.01345671$ &   $~~~0.03968709$    \\
			&              &                      &              &                      &                 &                      &       $0$       &         $0$          \\ \hline
	\end{tabular}}
\end{center}
\end{table*}
\textbf{Orthogonal wavelets}\label{APP_ortho_wavelet}\quad
Daubechies wavelets are orthogonal.
Table \ref{Tab_Daubechies_banks} shows their low-pass filter $\textbf{l} = \{l_k\}$ with order $p, 1\leq p\leq6$.
The length of the filter is $2p$.
The high-pass filter $\textbf{h} = \{h_k\}$ can be deduced from
\begin{equation}\label{eq_high_pass_bank}
h_k = (-1)^k l_{N-k},
\end{equation}
where $N$ is an odd number.
Daubechies(1) is Haar wavelet.

\textbf{Biorthogonal wavelets}\label{APP_bior_wavelet}\quad
Cohen wavelets are symmetric biorthogonal wavelets.
Each Cohen wavelet has four filters $\textbf{l}$, $\textbf{h}$, $\tilde{\textbf{l}}$, and $\tilde{\textbf{h}}$.
While a signal is decomposed using filters $\textbf{l}$ and $\textbf{h}$,
it can be reconstructed using the dual filters $\tilde{\textbf{l}}$ and $\tilde{\textbf{h}}$.
Cohen wavelet is with two order parameters $p$ and $\tilde{p}$.
Table \ref{Tab_CDF_banks} shows the low-pass filters with orders $2\leq p = \tilde{p}\leq5$.
Their high-pass filters can be deduced from
\begin{eqnarray}\label{eq_high_pass_bank_bior}
&h_k = (-1)^k \tilde{l}_{N-k},&\\
&\tilde{h}_k = (-1)^k l_{N-k},&
\end{eqnarray}
where $N$ is an odd number.
Cohen$(1,1)$ is Haar wavelet.

Our DWT and IDWT layers are applicable to any discrete orthogonal or biorthogonal wavelets.
With slight modifications, the layers could be applicable to other wavelet tools, 
such as multi-wavelets, ridgelet, curvelet, bandelet, contourlet, dual-tree complex wavelet, etc.

\subsection*{B. More example feature maps for WaveCNets}
In deep networks, the down-sampling operations ignoring the classic sampling theorem introduce at least two drawbacks:
breaking basic object structures and accumulating noise in the feature maps.
These drawbacks are related with the aliasing introduced by the down-sampling,
and we call them \textbf{aliasing effects} in deep networks.

\begin{figure*}[!bpt]
	\centering
	\subfigure[]
	{\includegraphics*[scale=0.25, viewport=175 65 1025 575]{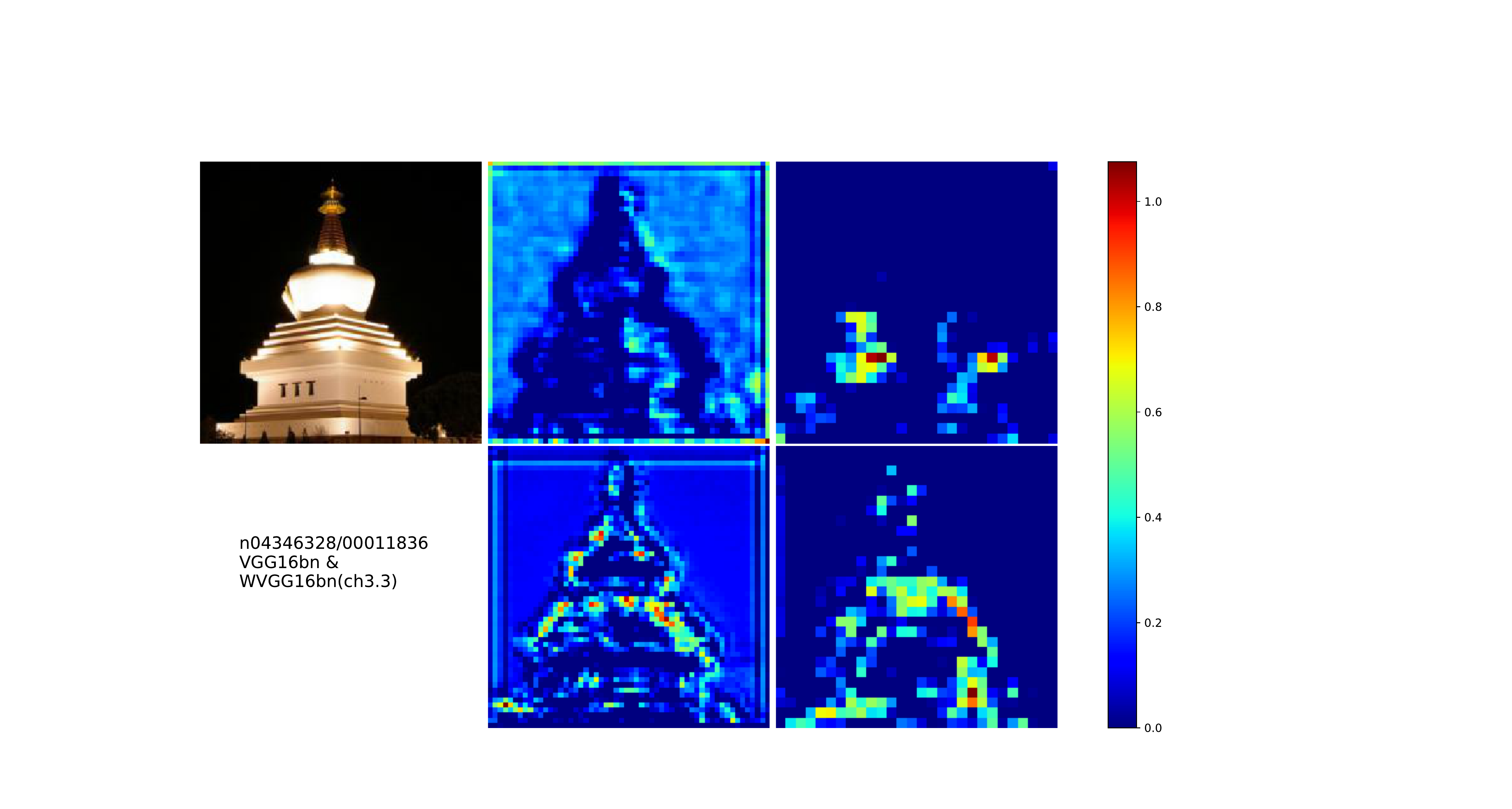}}\hspace{30pt}
	\subfigure[]
	{\includegraphics*[scale=0.25, viewport=175 65 1025 575]{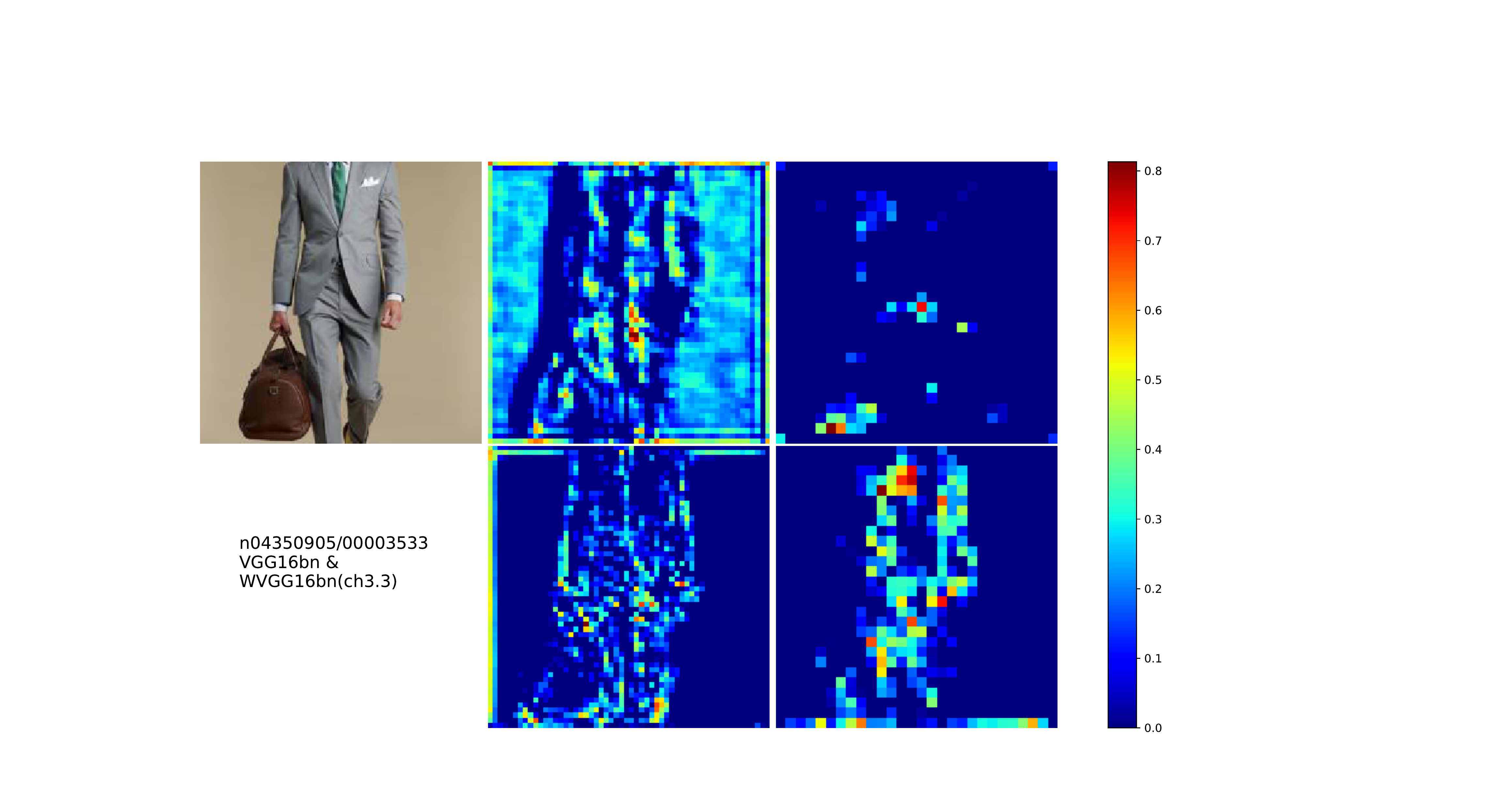}}\\
	\subfigure[]
	{\includegraphics*[scale=0.25, viewport=175 65 1025 575]{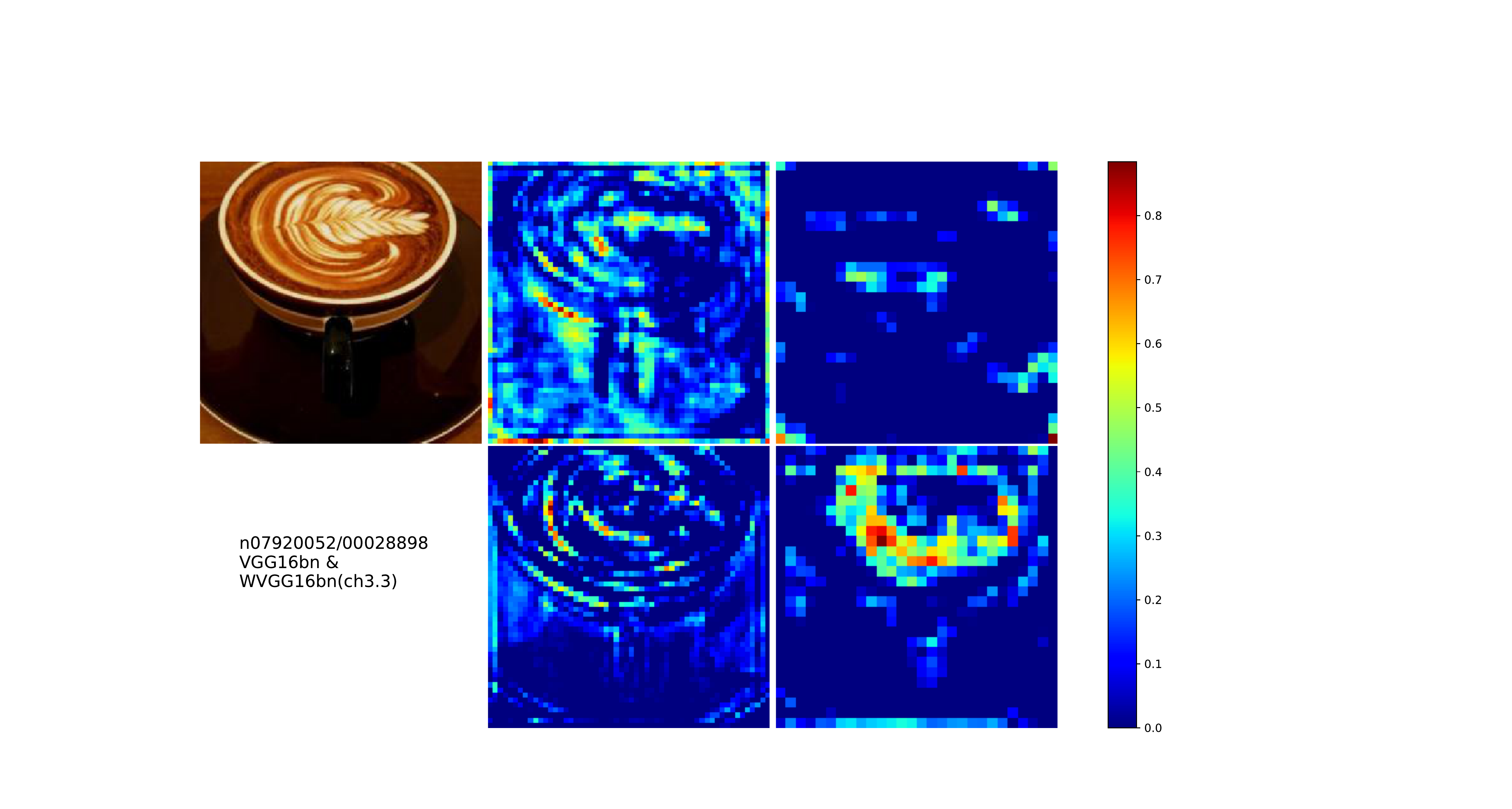}}\hspace{30pt}
	\subfigure[]
	{\includegraphics*[scale=0.25, viewport=175 65 1025 575]{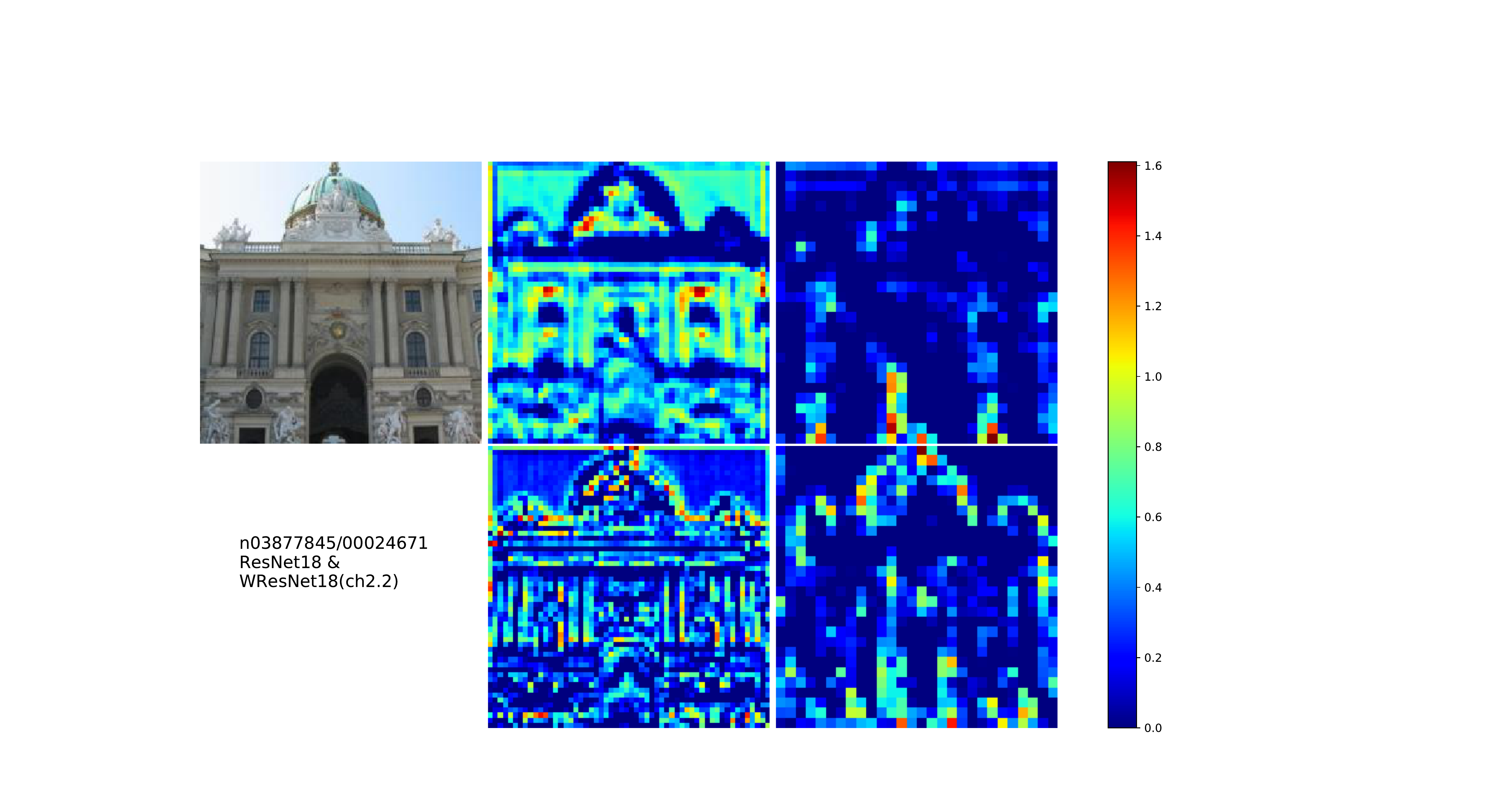}}\\
	\subfigure[]
	{\includegraphics*[scale=0.25, viewport=175 65 1025 575]{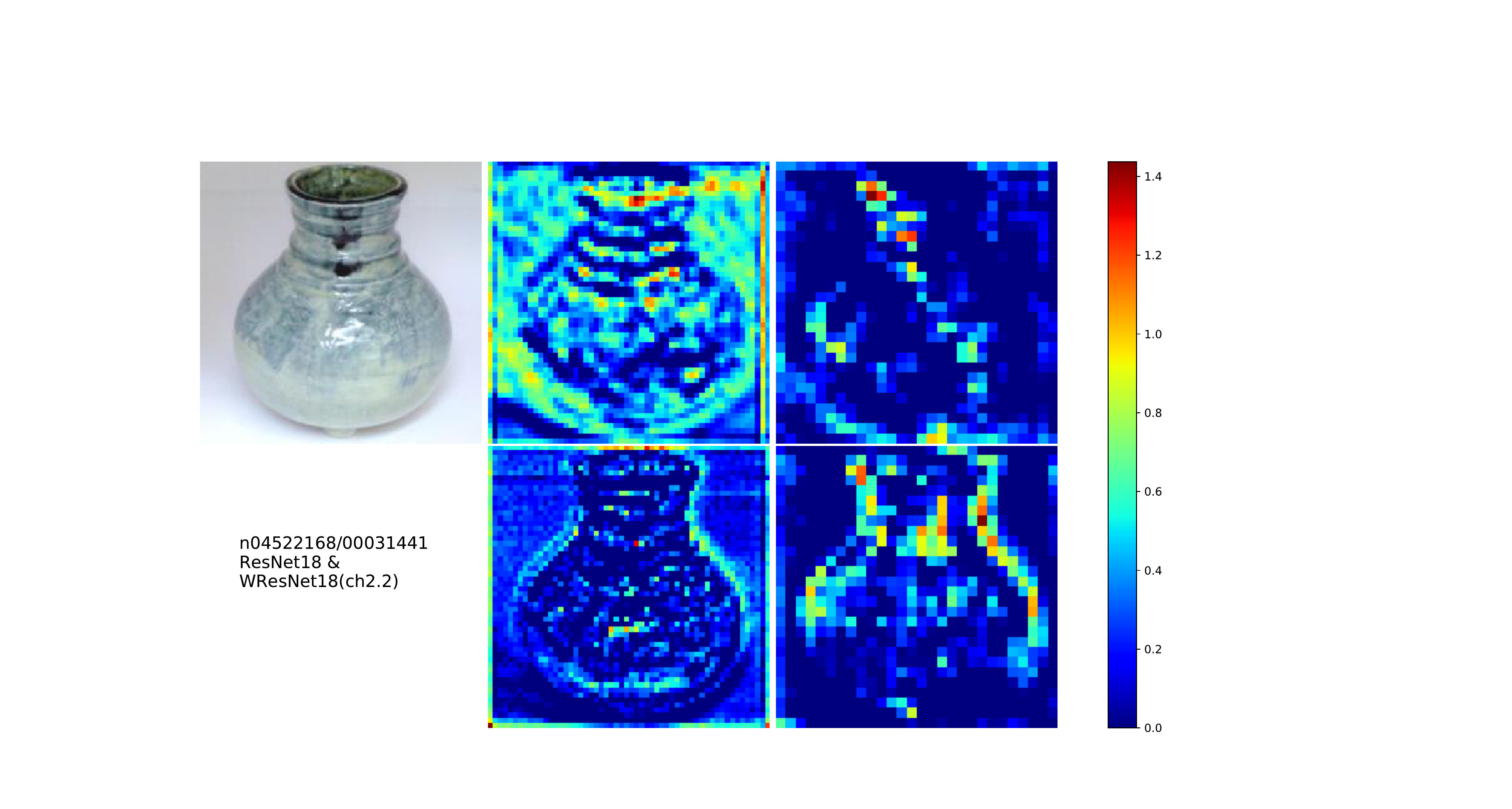}}\hspace{30pt}
	\subfigure[]
	{\includegraphics*[scale=0.25, viewport=175 65 1025 575]{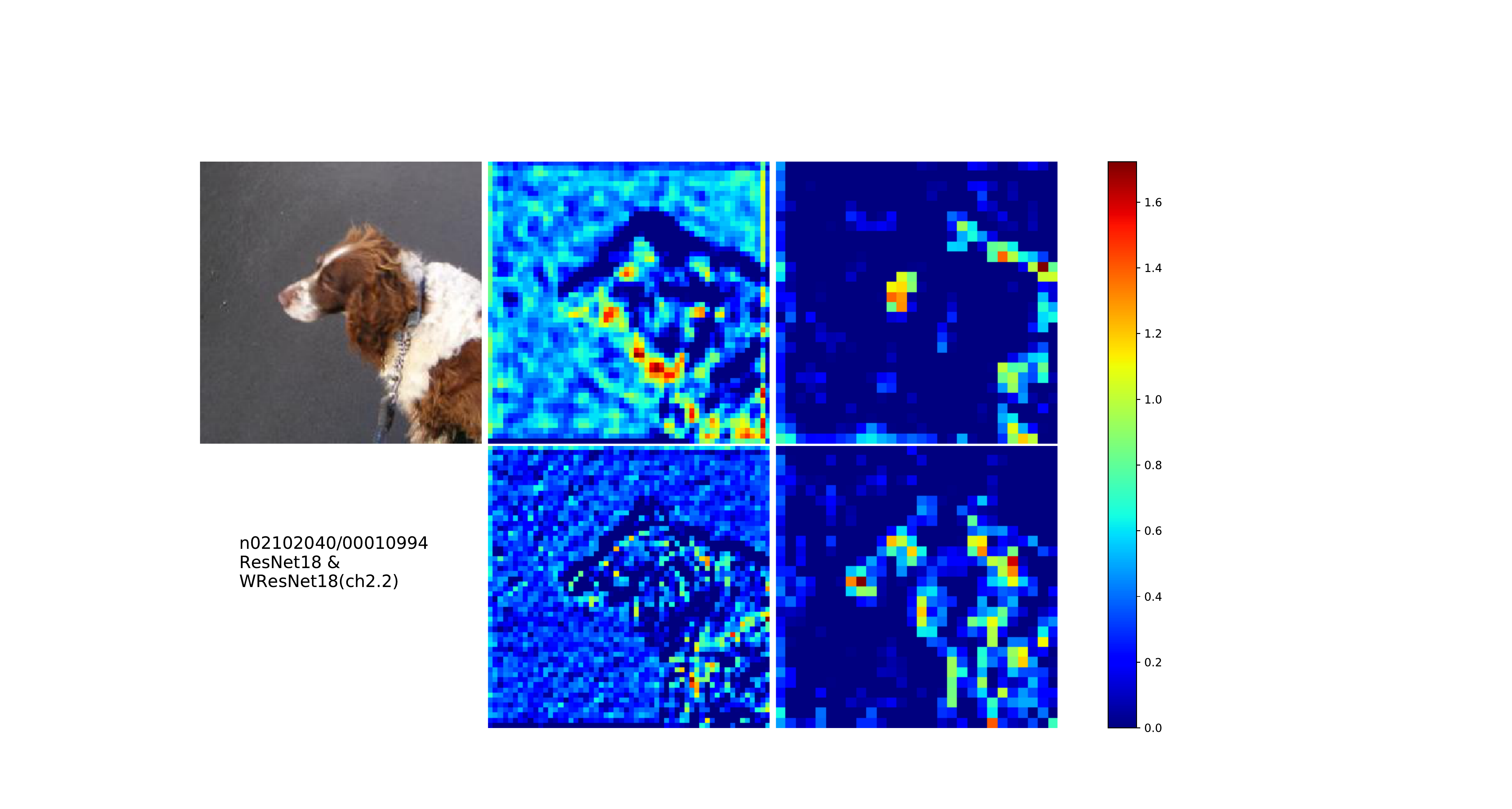}}\\
	\subfigure[]
	{\includegraphics*[scale=0.25, viewport=175 65 1025 575]{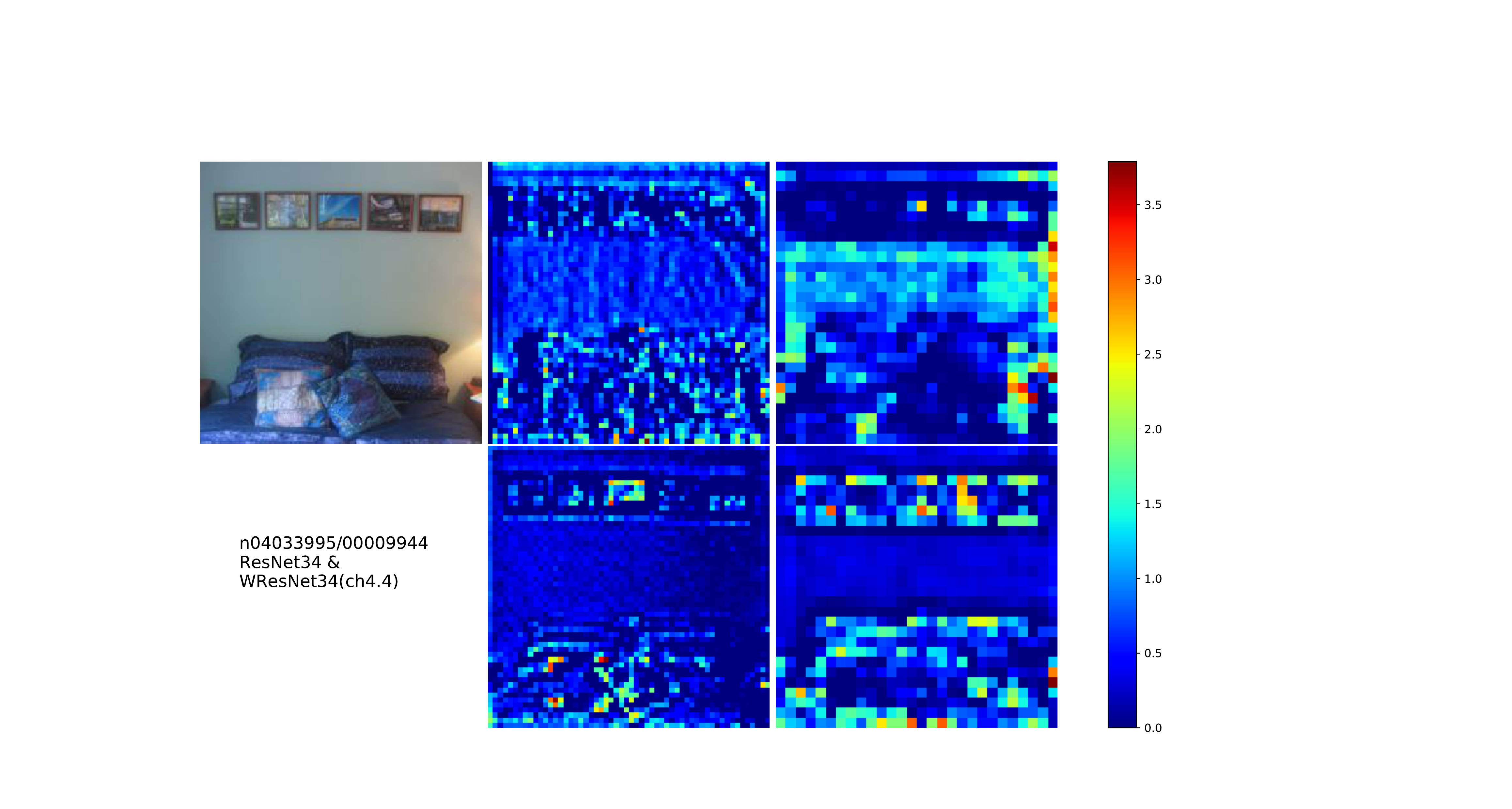}}\hspace{30pt}
	\subfigure[]
	{\includegraphics*[scale=0.25, viewport=175 65 1025 575]{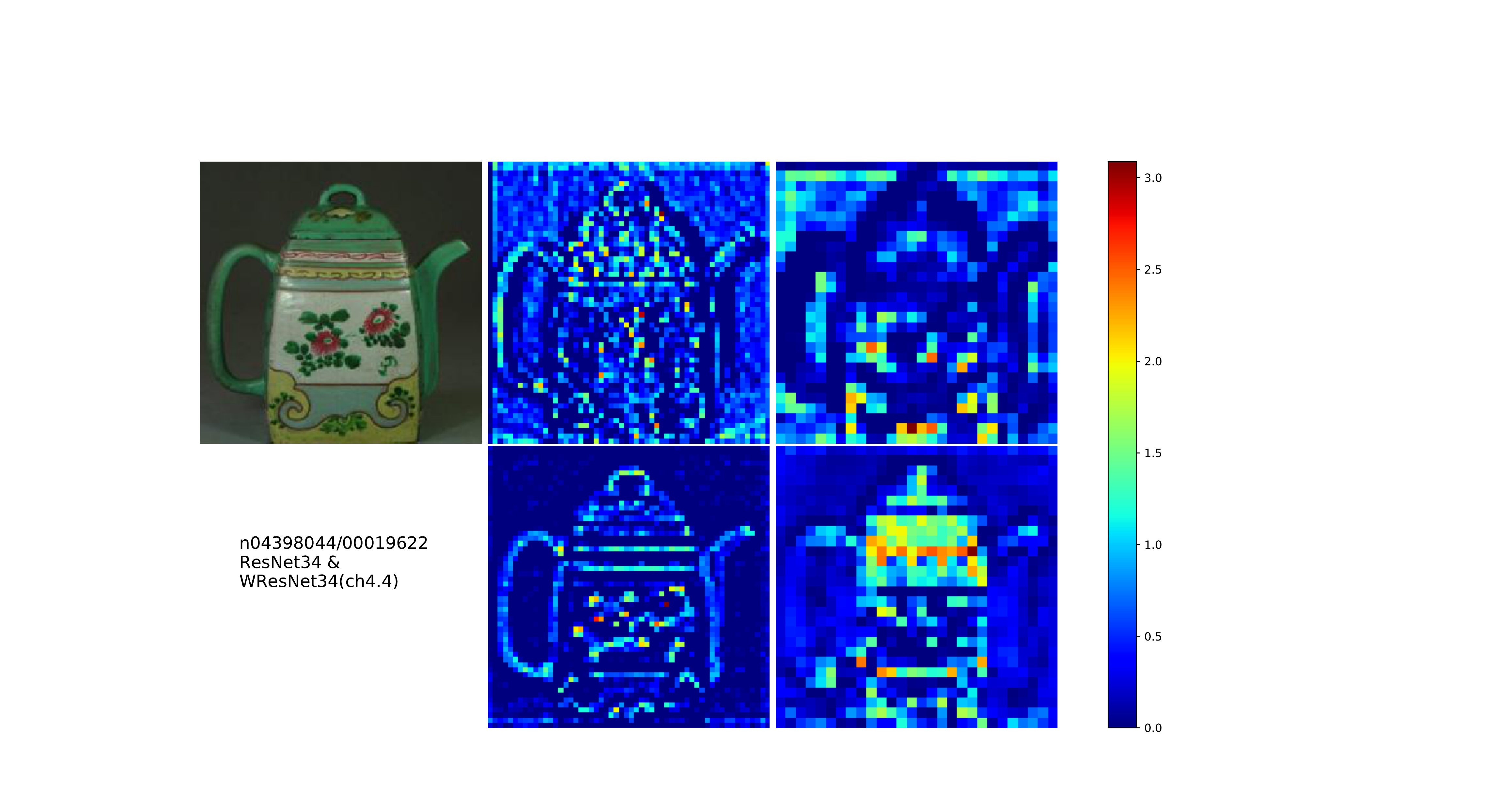}}
	\caption{The feature maps of CNNs (top) and WaveCNets (bottom).}
	\label{fig_feature_maps_more_0}
\end{figure*}

\begin{figure*}[!bpt]
	\centering
	\subfigure[]
	{\includegraphics*[scale=0.25, viewport=175 65 1025 575]{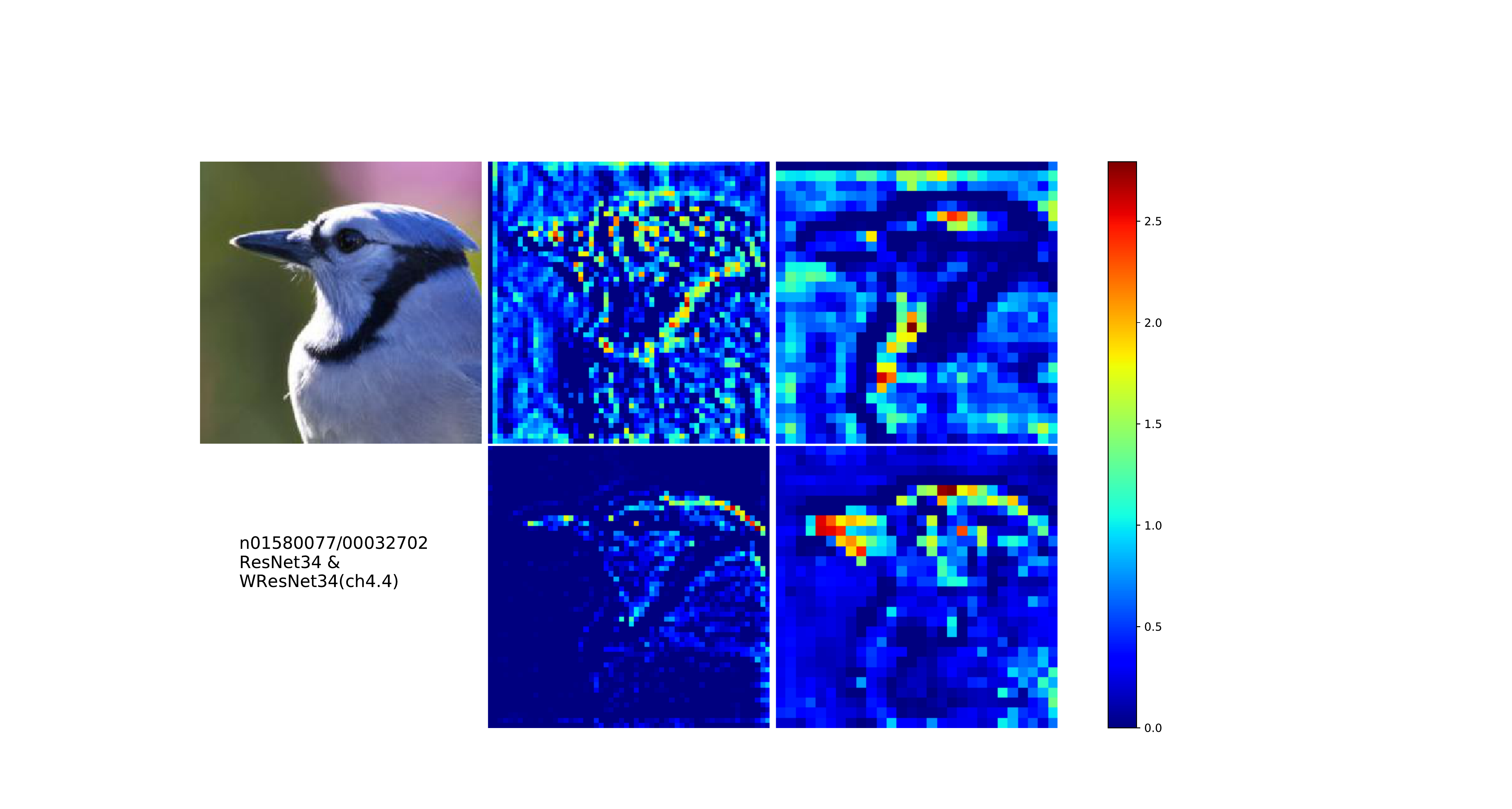}}\hspace{30pt}
	\subfigure[]
	{\includegraphics*[scale=0.25, viewport=175 65 1025 575]{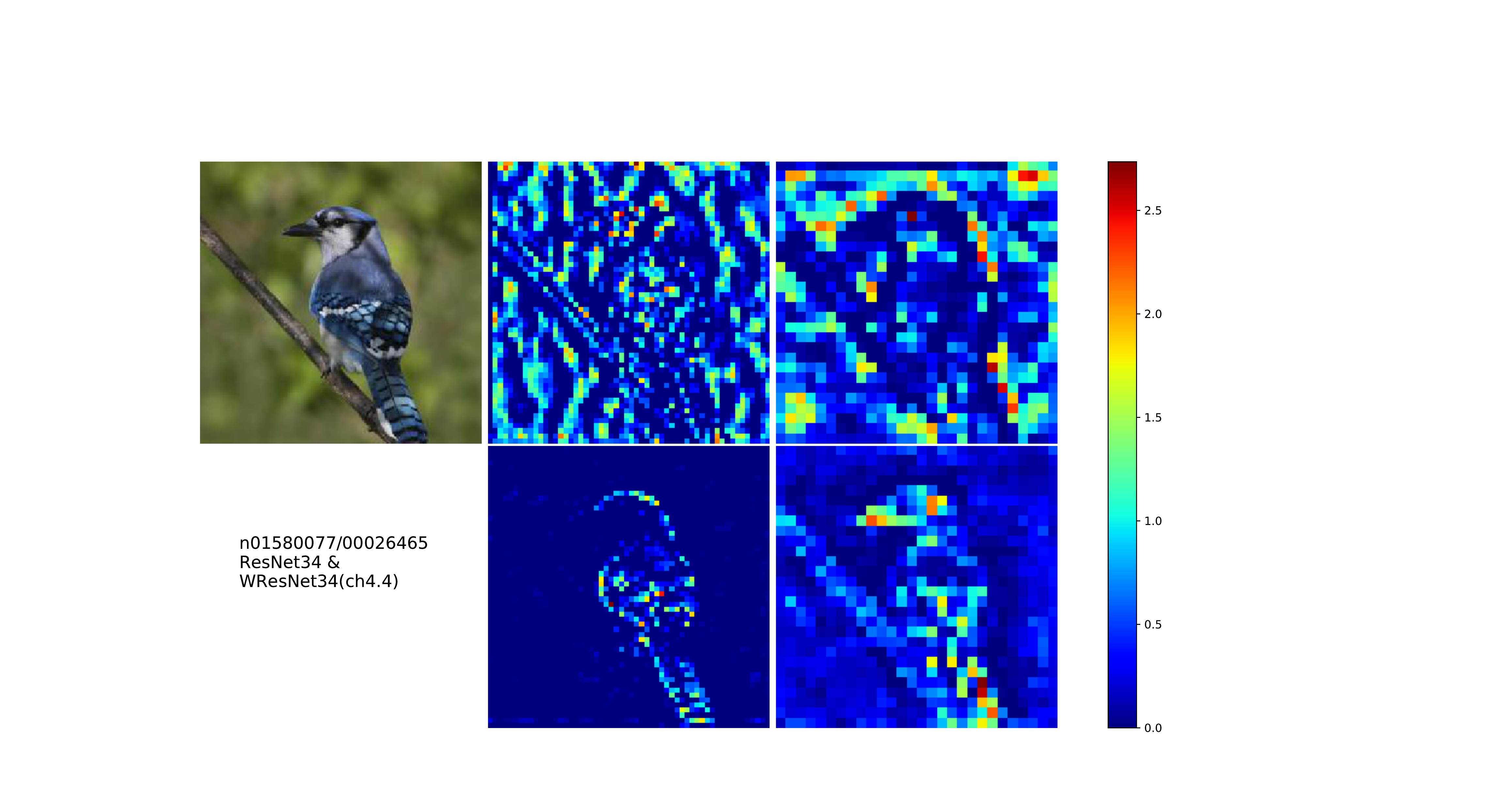}}\\
	\subfigure[]
	{\includegraphics*[scale=0.25, viewport=175 65 1025 575]{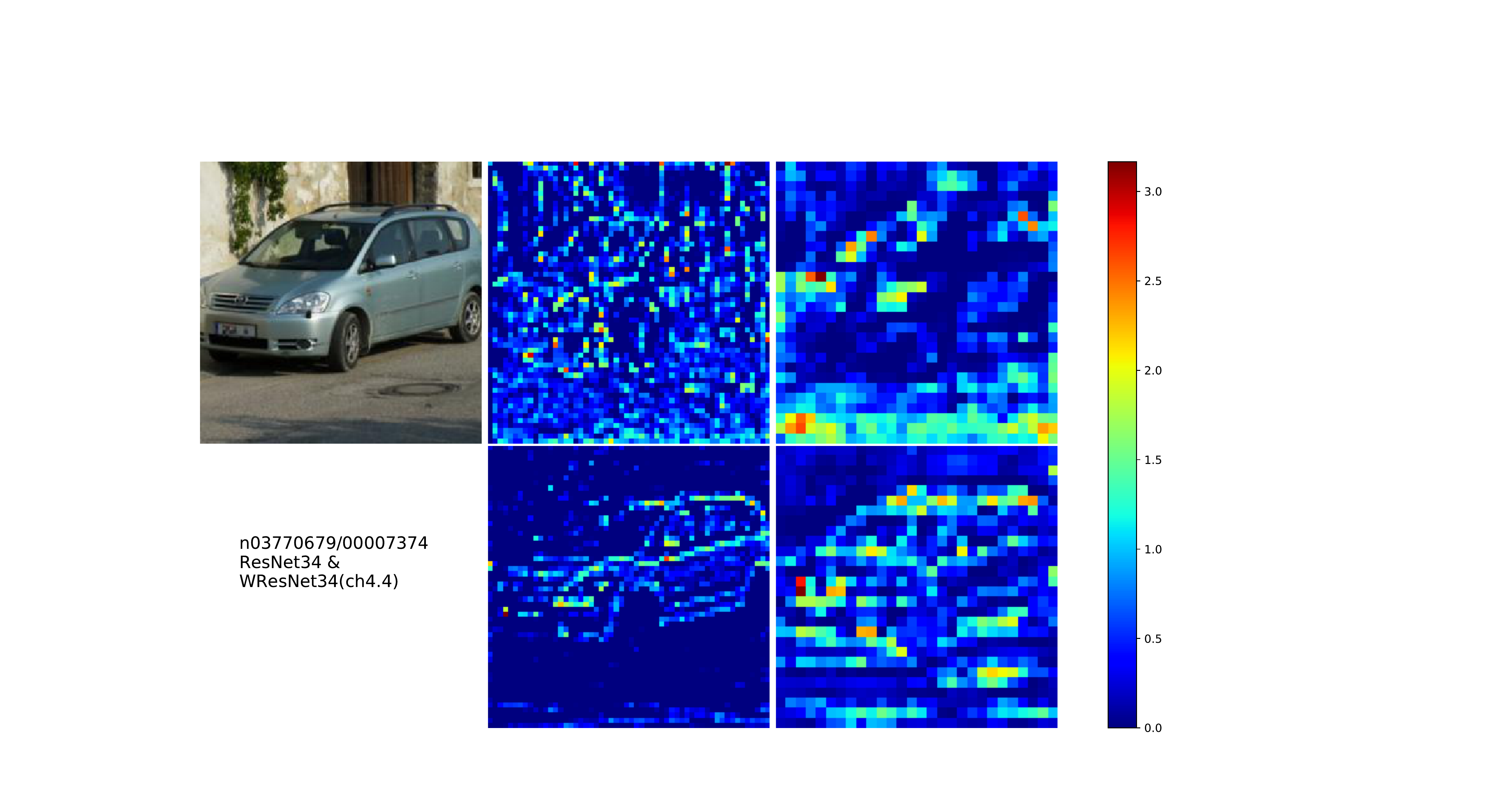}}\hspace{30pt}
	\subfigure[]
	{\includegraphics*[scale=0.25, viewport=175 65 1025 575]{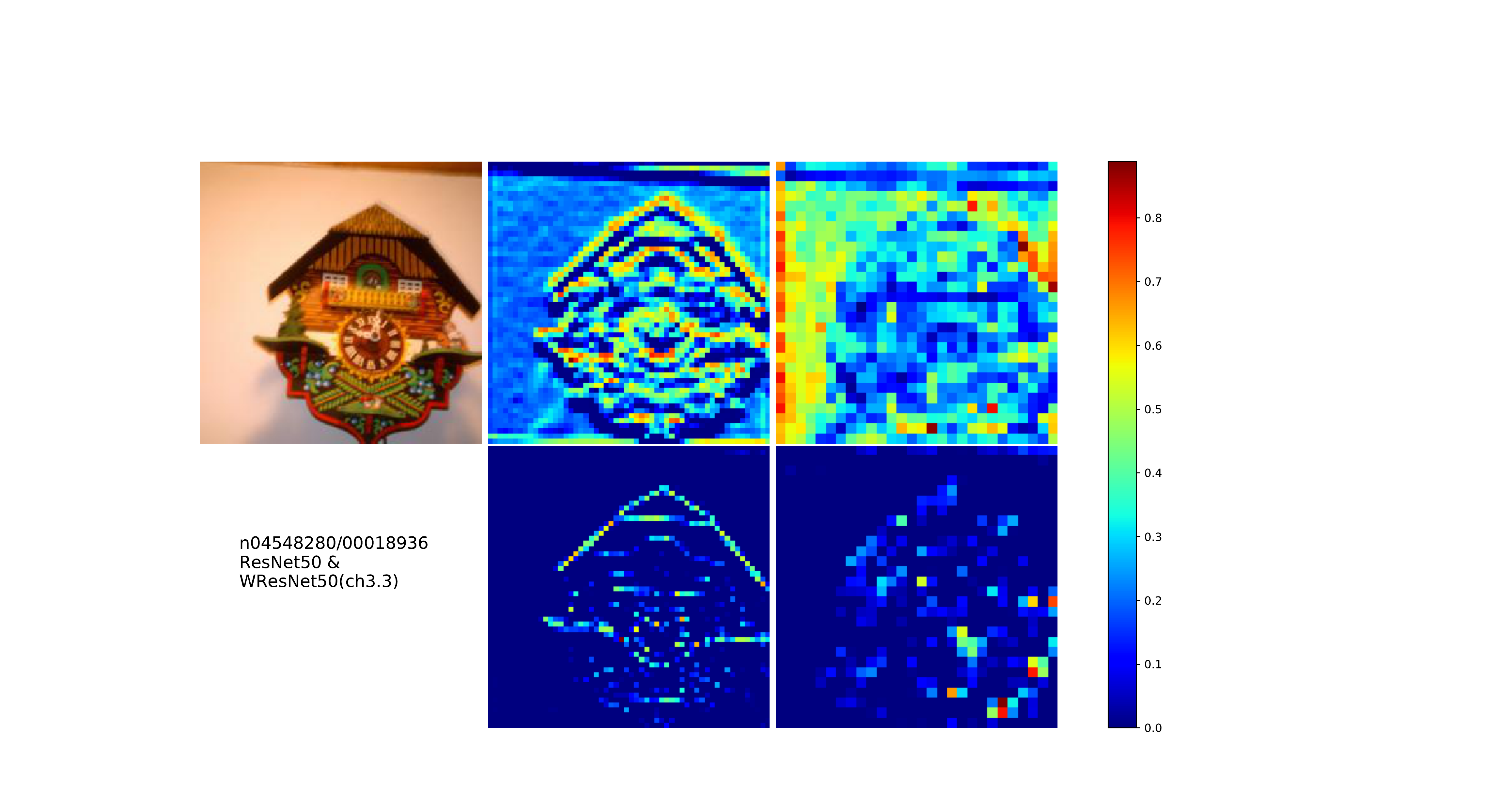}}\\
	\subfigure[]
	{\includegraphics*[scale=0.25, viewport=175 65 1025 575]{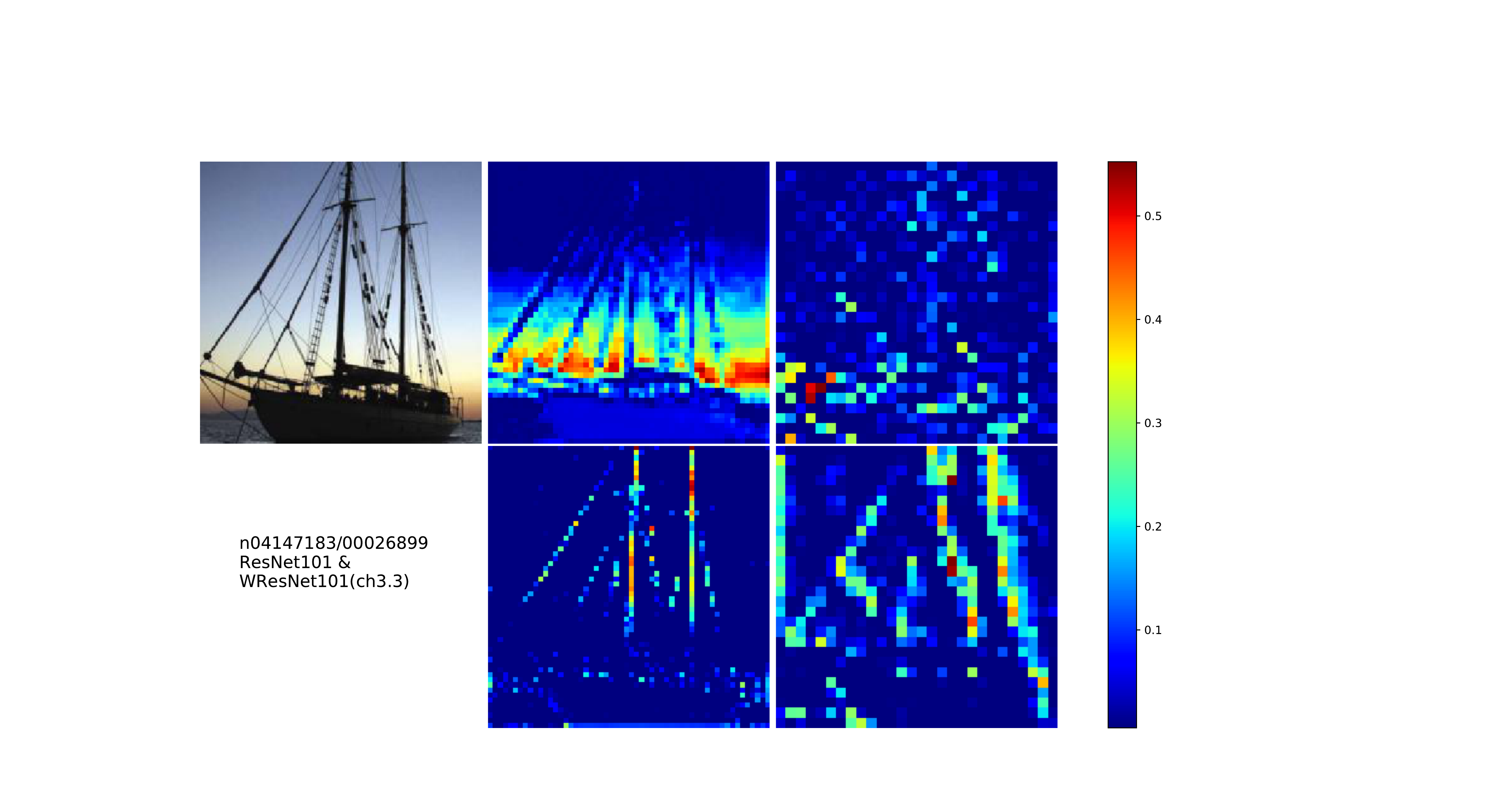}}\hspace{30pt}
	\subfigure[]
	{\includegraphics*[scale=0.25, viewport=175 65 1025 575]{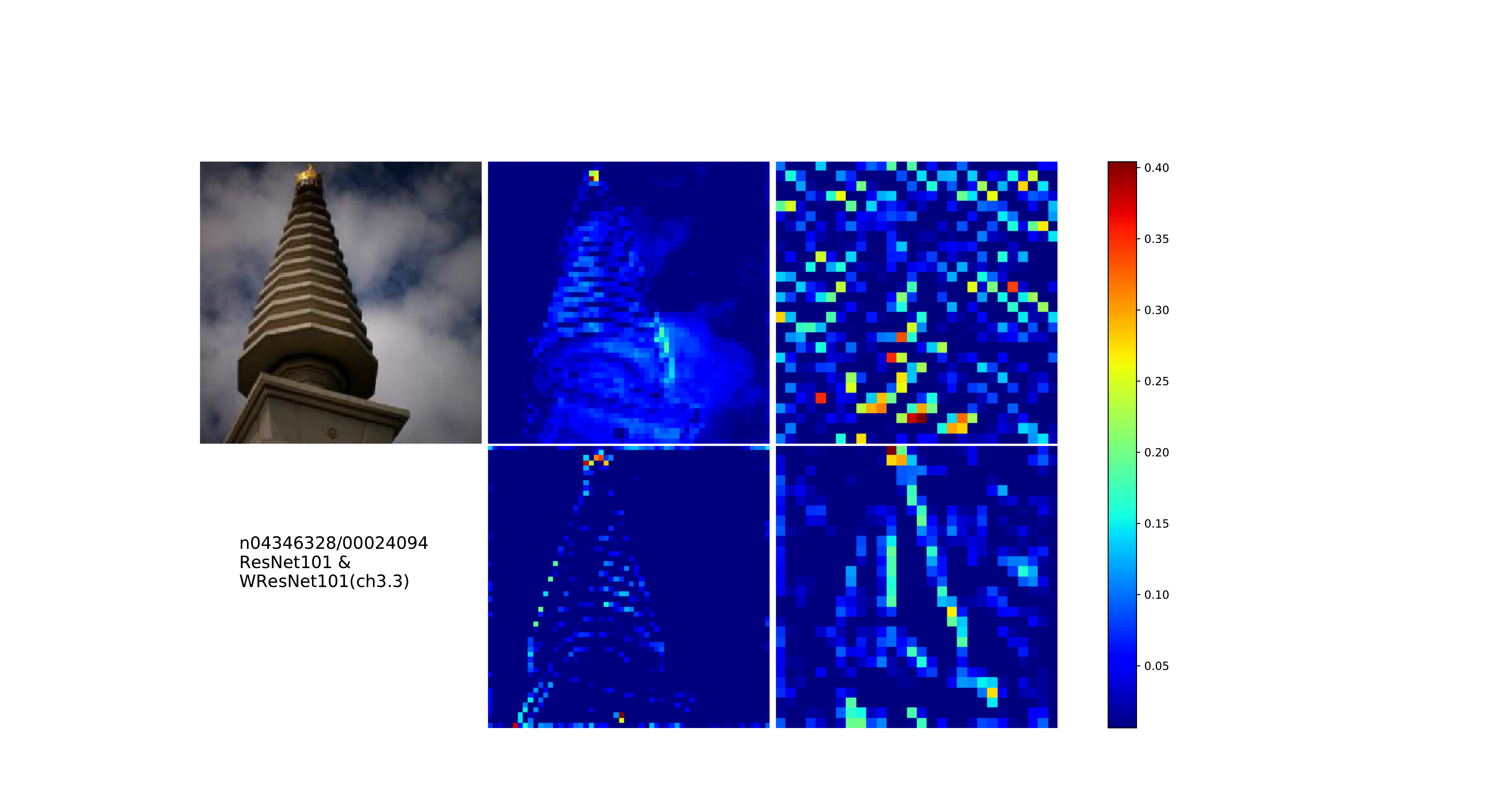}}\\
	\subfigure[]
	{\includegraphics*[scale=0.25, viewport=175 65 1025 575]{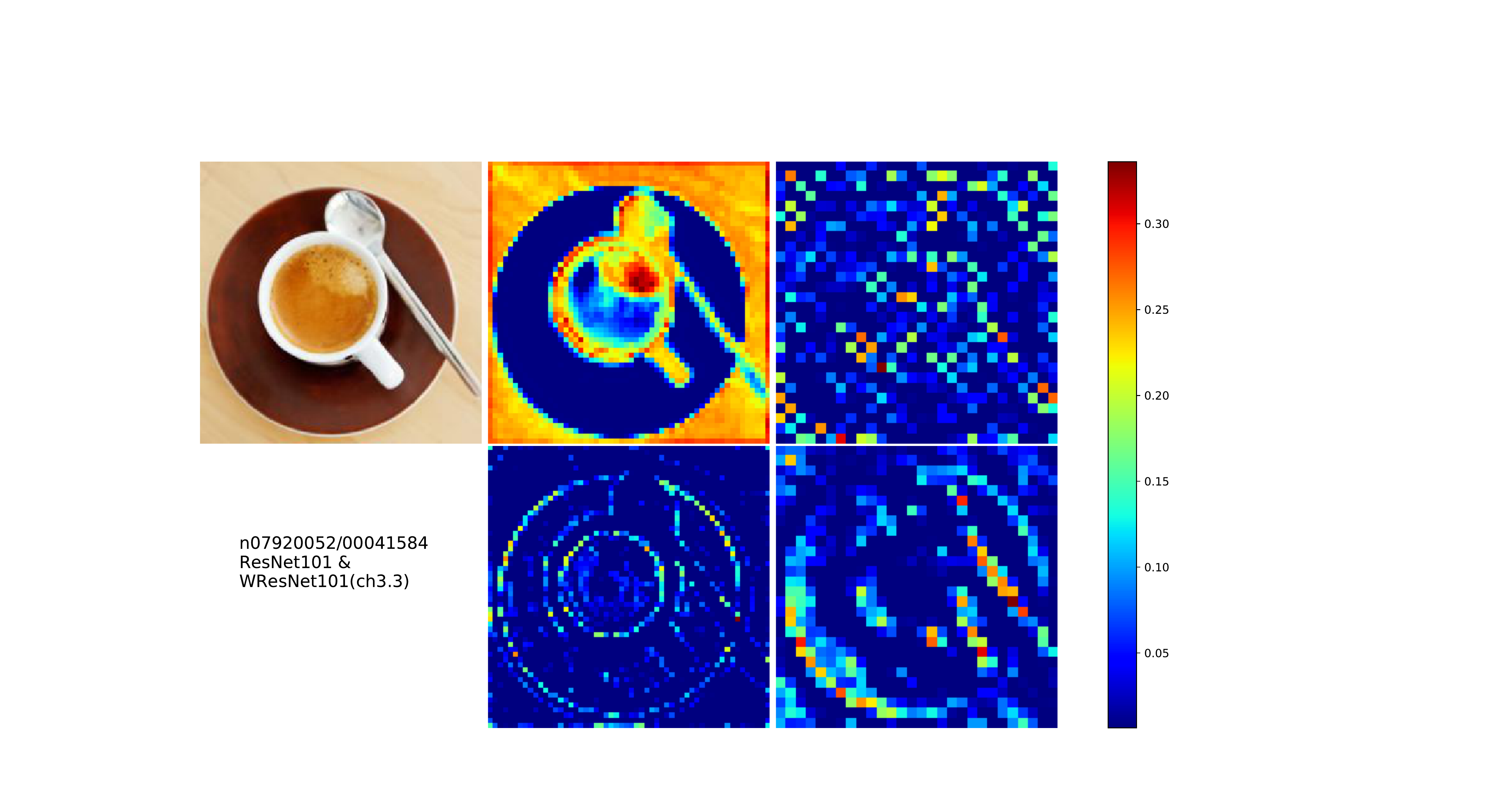}}\hspace{30pt}
	\subfigure[]
	{\includegraphics*[scale=0.25, viewport=175 65 1025 575]{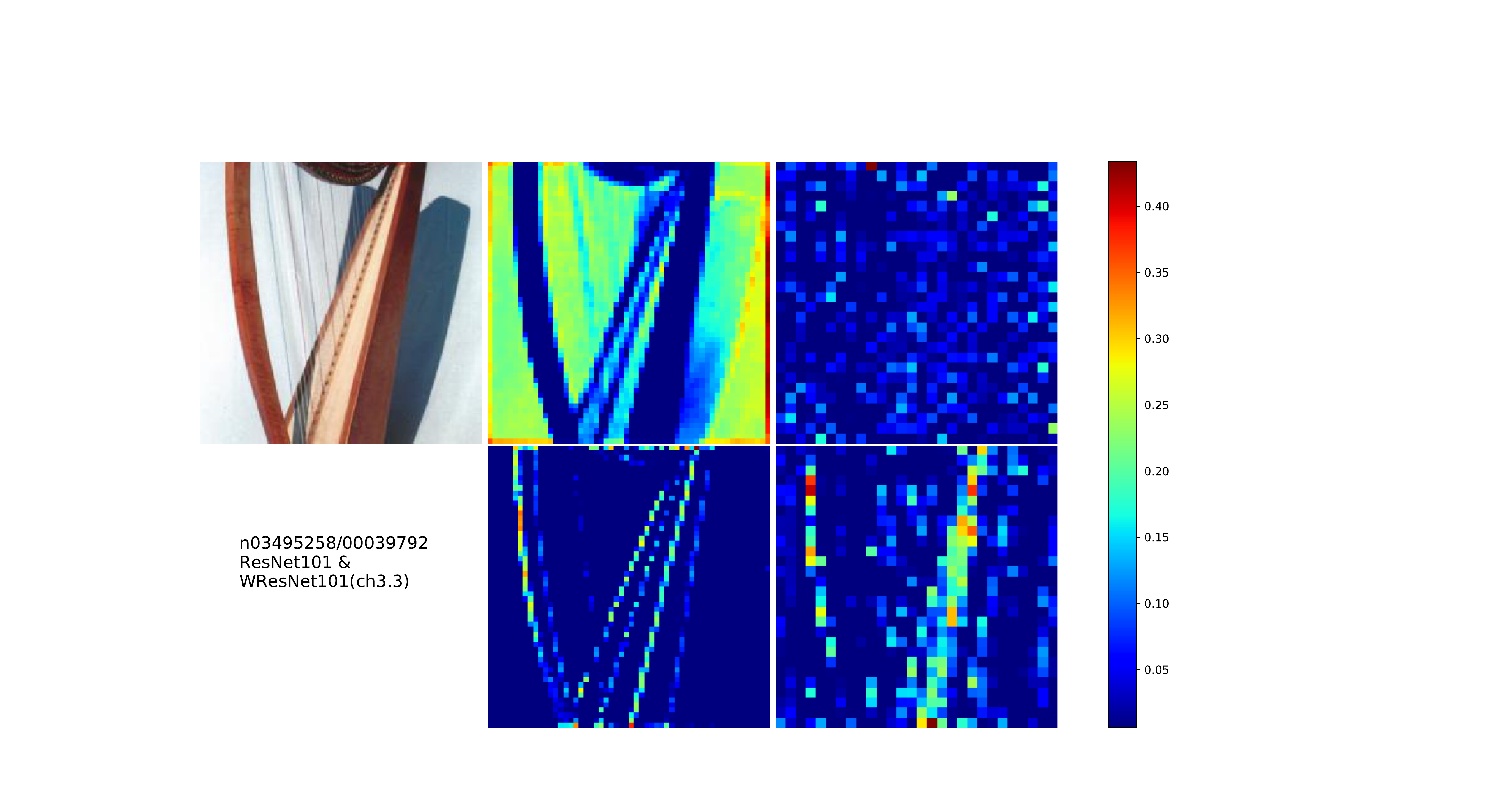}}
	\caption{The feature maps of CNNs (top) and WaveCNets (bottom).}
	\label{fig_feature_maps_more_1}
\end{figure*}

In WaveCNets, DWT is applied to maintain the basic object structures and resist the noise propagation,
for better accuracy and noise-robustness.
We have shown four example feature maps of well trained CNNs and WaveCNets in Fig. \ref{fig_feature_maps}.
To illustrate more persuasively the application of DWT in suppressing aliasing effects in deep networks,
Fig. \ref{fig_feature_maps_more_0} and Fig. \ref{fig_feature_maps_more_1} present more examples.


\subsection*{C. The detailed results on ImageNet-C}
\begin{table*}[!t]
	\scriptsize
	\caption{Corruption Error (CE) of WVGG16bn on ImageNet-C (lower is better).}\vspace{0pt}
	\label{Tab_mCE_VGG16bn}
	\begin{center}
	\setlength{\tabcolsep}{0.75mm}{
	\begin{tabular}{r|cccc|ccccc|ccccc|ccccc}\hline
		& \multicolumn{4}{c|}{Noise}& \multicolumn{5}{c|}{Blur}& \multicolumn{5}{c|}{Weather}& \multicolumn{5}{c}{Digital}\\\cline{2-20}
		& Gauss&Shot&Impulse&mCE&Defocus&Glass&Motion&Zoom&mCE&Snow&Frost&Fog&Bright&mCE&Contrast&Elastic&Pixel&Jpeg&mCE\\\hline
baseline&  86.28&   87.48&   89.11& 87.62&  83.95&   \textbf{94.80}&   86.36&   87.56& 88.17&  83.25&   \textbf{79.98}&   72.06&   \textbf{63.53}& 74.70&  75.02&   \textbf{95.22}&   94.87&   \textbf{88.88}& \textbf{88.50}\\\cdashline{1-20}[2.25pt/3pt]
haar&      85.80&   86.67&   89.82& 87.43&  84.24&   95.23&   85.46&   86.84& 87.94&  84.62&   80.68&   72.02&   64.38& 75.43&  75.74&   96.40&   96.49&   89.34& 89.49\\
ch2.2&     87.29&   87.80&   88.29& 87.79&  \textbf{83.63}&   95.63&   84.91&   \textbf{86.08}& \textbf{87.56}&  \textbf{82.70}&   80.65&   \textbf{71.55}&   63.64& \textbf{74.63}&  \textbf{74.66}&   96.17&   92.95&   90.34& 88.53\\
ch3.3&     86.40&   87.39&   91.04& 88.28&  84.03&   95.90&   \textbf{85.30}&   86.34& 87.89&  83.81&   81.01&   72.43&   64.14& 75.35&  75.18&   96.05&   93.34&   89.73& 88.58\\
ch4.4&     \textbf{85.01}&   \textbf{85.45}&   \textbf{86.44}& \textbf{85.63}&  84.34&   95.58&   85.77&   86.59& 88.07&  84.21&   82.11&   73.02&   64.73& 76.02&  76.62&   96.72&   \textbf{91.19}&   90.13& 88.67\\
ch5.5&     88.38&   89.14&   92.01& 89.84&  84.56&   96.80&   86.02&   86.77& 88.54&  86.58&   82.51&   74.24&   65.52& 77.21&  76.57&   97.50&   93.46&   90.23& 89.44\\\cdashline{1-20}[2.25pt/3pt]
db2&       86.54&   88.27&   87.51& 87.44&  84.46&   95.69&   85.43&   86.41& 88.00&  84.37&   81.46&   72.94&   64.76& 75.88&  76.57&   96.35&   93.14&   90.86& 89.23\\\hline
	\end{tabular}}
	\end{center}
\end{table*}
\begin{table*}[!t]
	\scriptsize
	\caption{Corruption Error (CE) of WResNet18 on ImageNet-C (lower is better).}\vspace{0pt}
	\label{Tab_mCE_ResNet18}
	\begin{center}
	\setlength{\tabcolsep}{0.75mm}{
	\begin{tabular}{r|cccc|ccccc|ccccc|ccccc}\hline
		& \multicolumn{4}{c|}{Noise}& \multicolumn{5}{c|}{Blur}& \multicolumn{5}{c|}{Weather}& \multicolumn{5}{c}{Digital}\\\cline{2-20}
		& Gauss&Shot&Impulse&mCE&Defocus&Glass&Motion&Zoom&mCE&Snow&Frost&Fog&Bright&mCE&Contrast&Elastic&Pixel&Jpeg&mCE\\\hline
baseline& 87.15&   88.47&   91.30&  88.97&   83.82&   91.43&   86.82&   88.70&   87.69&    86.10&   84.40&   78.48&   68.90&    79.47&   78.29&   90.23&   80.40&    \textbf{85.46}&    83.60\\\cdashline{1-20}[2.25pt/3pt]
haar&     80.64&   80.94&   81.16&  80.91&   80.18&   90.55&   84.04&   86.49&   85.32&    85.04&   81.93&   73.32&   65.78&    76.52&   75.72&   \textbf{87.78}&   \textbf{74.87}&    87.77  &  81.54  \\
ch2.2&    \textbf{80.15}&   \textbf{80.49}&   \textbf{80.50}&  \textbf{80.38}&   79.65&   \textbf{89.79}&   83.61&   84.82&   \textbf{84.47}&    84.91&   80.84&   73.99&   66.34&    76.52&   75.07&   88.19&   75.07&    88.61 &   \textbf{81.73} \\
ch3.3&    80.85&   81.44&   80.77&  81.02&   \textbf{79.28}&   91.20&   \textbf{82.71}&   85.52&   84.68&    84.48&   81.20&   \textbf{71.76}&   \textbf{65.44}&    \textbf{75.72}&   \textbf{73.77}&   89.66&   77.46&    86.06 &   81.74 \\
ch4.4&    81.83&   82.65&   82.10&  82.19&   79.55&   91.01&   82.88&   \textbf{84.55}&   84.50&    \textbf{83.91}&   \textbf{80.81}&   73.95&   66.27&    76.23&   75.67&   90.21&   78.35&    86.10 &   82.58 \\
ch5.5&    83.60&   83.87&   83.84&  83.77&   80.73&   91.64&   83.04&   85.45&   85.21&    84.39&   81.42&   73.83&   67.10&    76.68&   76.21&   91.07&   78.95&    89.35 &   83.89 \\\cdashline{1-20}[2.25pt/3pt]
db2&      82.30&   82.65&   82.68&  82.54&   80.16&   91.22&   83.55&   84.73&   84.92&    85.42&   81.74&   74.34&   66.51&    77.00&   75.92&   90.41&   79.54&    91.71   & 84.39   \\
db3&      83.75&   84.14&   83.87&  83.92&   81.36&   90.92&   84.14&   86.24&   85.66&    85.01&   81.73&   75.79&   68.14&    77.67&   78.47&   90.02&   78.41&    89.35   & 84.06   \\
db4&      86.00&   85.83&   86.85&  86.22&   82.12&   92.93&   85.75&   87.38&   87.04&    85.31&   83.02&   77.87&   68.97&    78.79&   79.33&   91.07&   74.95&    85.63   & 82.74   \\
db5&      85.22&   85.86&   85.33&  85.47&   82.96&   92.65&   87.83&   88.63&   88.02&    87.60&   85.21&   78.66&   71.33&    80.70&   80.85&   91.11&   78.21&    92.99   & 85.79   \\
db6&      86.29&   86.73&   86.68&  86.57&   84.72&   94.50&   87.73&   88.68&   88.91&    87.88&   86.62&   80.58&   72.81&    81.97&   82.29&   93.46&   78.32&    95.53    & 87.40    \\\hline
	\end{tabular}}
	\end{center}
\end{table*}
\begin{table*}[!t]
	\scriptsize
	\caption{Corruption Error (CE) of WResNet34 on ImageNet-C (lower is better).}\vspace{0pt}
	\label{Tab_mCE_ResNet34}
	\begin{center}
	\setlength{\tabcolsep}{0.75mm}{
	\begin{tabular}{r|cccc|ccccc|ccccc|ccccc}\hline
		& \multicolumn{4}{c|}{Noise}& \multicolumn{5}{c|}{Blur}& \multicolumn{5}{c|}{Weather}& \multicolumn{5}{c}{Digital}\\\cline{2-20}
		& Gauss&Shot&Impulse&mCE&Defocus&Glass&Motion&Zoom&mCE&Snow&Frost&Fog&Bright&mCE&Contrast&Elastic&Pixel&Jpeg&mCE\\\hline
baseline&   81.36&   83.01&   84.94&   83.10&   76.04&   86.95&   79.59&   84.56&   81.79&   79.87&   77.02&   69.34&   61.97&   72.05&   71.80&   86.17&   70.54&   \textbf{74.60}& 75.78\\\cdashline{1-20}[2.25pt/3pt]
haar&       75.80&   77.21&   76.89&   76.64&   76.17&   87.25&   76.86&   81.08&   80.34&   80.51&   75.85&   68.33&   60.29&   71.25&   71.17&   84.10&   71.61&   80.17& 76.76\\
ch2.2&      76.91&   77.56&   78.36&   77.61&   73.49&   87.09&   75.77&   80.34&   79.17&   79.59&   \textbf{75.71}&   67.25&   59.59&   70.53&   69.87&   86.27&   67.28&   77.33& 75.19\\
ch3.3&      73.73&   74.66&   74.50&   74.30&   74.42&   89.02&   76.15&   \textbf{79.83}&   79.86&   79.59&   76.74&   \textbf{65.70}&   \textbf{58.82}&   \textbf{70.21}&   \textbf{68.95}&   84.51&   71.95&   77.47& 75.72\\
ch4.4&      74.60&   76.21&   77.75&   76.19&   \textbf{72.99}&   88.37&   \textbf{73.25}&   80.17&   \textbf{78.69}&   79.51&   76.01&   67.48&   60.18&   70.80&   69.93&   85.46&   65.42&   77.37& 74.54\\
ch5.5&      75.92&   76.68&   75.41&   76.00&   73.60&   87.49&   75.10&   81.23&   79.36&   \textbf{78.80}&   75.99&   69.01&   60.41&   71.05&   71.39&   85.86&   67.06&   77.14& 75.36\\\cdashline{1-20}[2.25pt/3pt]
db2&        \textbf{71.80}&   \textbf{73.24}&   73.16&   \textbf{72.73}&   73.82&   \textbf{86.79}&   75.68&   81.25&   79.38&   80.77&   76.34&   67.99&   60.28&   71.34&   70.41&   \textbf{83.63}&   65.64&   76.41& \textbf{74.02}\\
db3&        75.77&   76.78&   76.74&   76.43&   73.77&   88.69&   76.30&   81.03&   79.94&   80.67&   77.12&   70.22&   60.68&   72.17&   72.34&   85.61&   \textbf{64.53}&   81.27& 75.94\\
db4&        78.21&   79.46&   79.19&   78.96&   73.51&   88.68&   76.80&   81.70&   80.17&   79.16&   77.38&   69.90&   61.79&   72.06&   71.75&   84.77&   73.90&   79.91& 77.58\\
db5&        72.82&   73.58&   \textbf{73.06}&   73.15&   75.88&   88.75&   80.86&   84.83&   82.58&   80.36&   77.29&   70.71&   62.33&   72.67&   72.48&   85.18&   66.46&   80.59& 76.18\\
db6&        80.91&   81.98&   83.17&   82.02&   76.35&   91.06&   81.37&   83.92&   83.18&   82.25&   79.69&   70.77&   63.35&   74.02&   72.93&   87.84&   75.06&   88.55& 81.09\\\hline
	\end{tabular}}
	\end{center}
\end{table*}
\begin{table*}[!t]
	\scriptsize
	\caption{Corruption Error (CE) of WResNet50 on ImageNet-C (lower is better).}\vspace{0pt}
	\label{Tab_mCE_ResNet50}
	\begin{center}
	\setlength{\tabcolsep}{0.75mm}{
	\begin{tabular}{r|cccc|ccccc|ccccc|ccccc}\hline
		& \multicolumn{4}{c|}{Noise}& \multicolumn{5}{c|}{Blur}& \multicolumn{5}{c|}{Weather}& \multicolumn{5}{c}{Digital}\\\cline{2-20}
		& Gauss&Shot&Impulse&mCE&Defocus&Glass&Motion&Zoom&mCE&Snow&Frost&Fog&Bright&mCE&Contrast&Elastic&Pixel&Jpeg&mCE\\\hline
baseline&  79.77&   81.57&   82.58&   81.31&   74.71&   88.61&   78.03&   79.86&   80.30&    \textbf{77.83}&   74.84&   \textbf{66.11}&   \textbf{56.64}&   \textbf{68.85}&   71.43&   84.75&   76.92&   76.82&  77.48\\\cdashline{1-20}[2.25pt/3pt]
haar&      77.41&   78.80&   79.22&   78.48&   71.00&   86.38&   76.80&   77.18&   77.84&    80.22&   74.73&   66.18&   57.23&   69.59&   \textbf{70.90}&   84.06&   75.07&   76.51&  76.64\\
ch2.2&     76.08&   77.24&   77.28&   76.87&   71.92&   85.95&   77.54&   77.26&   78.17&    79.24&   75.02&   69.26&   58.44&   70.49&   72.25&   84.63&   68.18&   75.86&  75.23\\
ch3.3&     74.17&   75.63&   75.31&   75.04&   71.67&   86.49&   77.74&   77.89&   78.45&    80.67&   75.50&   66.44&   57.99&   70.15&   71.34&   84.87&   65.61&   \textbf{74.01}&  73.96\\
ch4.4&     76.09&   77.38&   76.72&   76.73&   72.00&   87.03&   77.62&   78.47&   78.78&    79.13&   75.30&   68.80&   57.31&   70.13&   72.03&   85.27&   71.17&   75.80&  76.07\\
ch5.5&     \textbf{71.49}&   \textbf{73.04}&   \textbf{71.70}&   \textbf{72.08}&   72.77&   86.09&   77.61&   77.45&   78.48&    78.55&   \textbf{74.00}&   67.82&   57.48&   69.46&   71.94&   85.99&   74.05&   75.88&  76.97\\\cdashline{1-20}[2.25pt/3pt]
db2&       78.64&   79.23&   78.51&   78.80&   \textbf{69.15}&   \textbf{85.08}&   \textbf{74.56}&   \textbf{76.80}&   \textbf{76.40}&    79.15&   74.96&   68.01&   57.76&   69.97&   71.05&   \textbf{82.56}&   \textbf{60.67}&   78.54&  \textbf{73.20}\\
db3&       78.74&   79.93&   79.36&   79.34&   70.71&   86.92&   76.26&   78.45&   78.09&    78.32&   76.12&   67.07&   57.44&   69.74&   72.04&   87.17&   70.70&   82.13&  78.01\\
db4&       77.16&   78.62&   78.19&   77.99&   73.74&   88.97&   80.20&   79.88&   80.70&    79.74&   76.18&   69.52&   59.29&   71.18&   73.49&   87.84&   74.10&   75.27&  77.67\\
db5&       77.46&   78.80&   78.74&   78.33&   75.36&   89.05&   78.68&   81.08&   81.04&    81.12&   77.97&   71.87&   61.07&   73.01&   74.26&   85.81&   73.96&   84.13&  79.54\\
db6&       79.23&   79.87&   80.23&   79.78&   77.25&   90.20&   82.20&   82.36&   83.00&    82.76&   79.86&   74.50&   64.29&   75.35&   77.66&   88.45&   81.67&   86.14&  83.48\\\hline
	\end{tabular}}
	\end{center}
\end{table*}
\begin{table*}[!t]
	\scriptsize
	\caption{Corruption Error (CE) of WResNet101 on ImageNet-C (lower is better).}\vspace{0pt}
	\label{Tab_mCE_ResNet101}
	\begin{center}
	\setlength{\tabcolsep}{0.75mm}{
	\begin{tabular}{r|cccc|ccccc|ccccc|ccccc}\hline
		& \multicolumn{4}{c|}{Noise}& \multicolumn{5}{c|}{Blur}& \multicolumn{5}{c|}{Weather}& \multicolumn{5}{c}{Digital}\\\cline{2-20}
		& Gauss&Shot&Impulse&mCE&Defocus&Glass&Motion&Zoom&mCE&Snow&Frost&Fog&Bright&mCE&Contrast&Elastic&Pixel&Jpeg&mCE\\\hline
baseline&  73.59&   75.31&   76.93&   75.28&   67.98&   81.58&   70.86&   74.17&    73.65&    73.26&   70.50&   62.07&   53.40& 64.81&  66.83&   77.23&   64.76&   67.11&  68.98\\\cdashline{1-20}[2.25pt/3pt]
haar&      70.33&   70.95&   70.10&   70.46&   \textbf{62.93}&   80.93&   \textbf{64.83}&   72.97&    \textbf{70.41}&    71.54&   67.27&   59.94&   50.46& 62.30&  \textbf{62.23}&   76.24&   57.39&   68.25&  66.03\\
ch2.2&     \textbf{64.25}&   \textbf{65.91}&   \textbf{65.04}&   \textbf{65.06}&   63.57&   \textbf{80.35}&   68.22&   \textbf{71.75}&    70.97&    71.55&   67.45&   60.21&   50.13& 62.33&  62.89&   \textbf{74.64}&   \textbf{52.33}&   67.96&  \textbf{64.46}\\
ch3.3&     69.32&   70.91&   69.24&   69.82&   63.78&   81.02&   71.07&   72.00&    71.97&    71.46&   68.60&   60.73&   51.36& 63.04&  64.02&   78.16&   59.05&   \textbf{64.48}&  66.42\\
ch4.4&     67.70&   69.30&   69.71&   68.91&   65.24&   81.05&   69.61&   72.36&    72.07&    71.75&   \textbf{67.18}&   60.10&   50.64& 62.41&  63.47&   77.76&   63.64&   66.96&  67.96\\
ch5.5&     69.67&   71.00&   70.21&   70.30&   64.22&   82.00&   70.15&   74.12&    72.62&    \textbf{71.10}&   67.21&   \textbf{59.07}&   \textbf{50.07}& \textbf{61.86}&  62.31&   78.53&   59.16&   65.01&  66.25\\\cdashline{1-20}[2.25pt/3pt]
db2&       69.70&   71.33&   71.25&   70.76&   65.24&   81.50&   73.36&   74.26&    73.59&    73.66&   68.56&   61.58&   50.65& 63.61&  64.40&   75.81&   62.48&   69.27&  67.99\\  \hline
	\end{tabular}}
	\end{center}
\end{table*}
\begin{table*}[!t]
	\scriptsize
	\caption{Corruption Error (CE) of WDenseNet121 on ImageNet-C (lower is better).}
	\label{Tab_mCE_DenseNet121}
	\begin{center}
	\setlength{\tabcolsep}{0.75mm}{
	\begin{tabular}{r|cccc|ccccc|ccccc|ccccc}\hline
		& \multicolumn{4}{c|}{Noise}& \multicolumn{5}{c|}{Blur}& \multicolumn{5}{c|}{Weather}& \multicolumn{5}{c}{Digital}\\\cline{2-20}
		& Gauss&Shot&Impulse&mCE&Defocus&Glass&Motion&Zoom&mCE&Snow&Frost&Fog&Bright&mCE&Contrast&Elastic&Pixel&Jpeg&mCE\\\hline
baseline&80.79&   82.34&   83.75&   82.29&   76.82&   88.72&   80.54&   82.58&   82.16&  78.24&   74.86&   70.18&   59.50&   70.70&  74.07&   87.39&   74.47&   \textbf{74.57}&   77.62\\\cdashline{1-20}[2.25pt/3pt]
haar&	 77.35&   78.74&   78.91&   78.33&   72.82&   86.99&   79.46&   80.04&   79.83&  78.64&   74.61&   67.17&   58.44&   69.72&  70.85&   \textbf{84.16}&   75.72&   77.12&   76.96\\
ch2.2&	 76.15&   77.07&   77.69&   76.97&   72.07&   87.21&   78.04&   80.84&   79.54&  79.85&   73.58&   66.23&   \textbf{57.60}&   69.32&  69.54&   86.15&   74.56&   77.34&   76.90\\
ch3.3&	 77.14&   77.65&   80.15&   78.31&   73.97&   88.14&   78.77&   80.65&   80.38&  77.78&   74.63&   67.03&   58.50&   69.48&  72.12&   85.84&   72.26&   75.44&   \textbf{76.42}\\
ch4.4&	 \textbf{75.26}&   \textbf{76.38}&   \textbf{76.39}&   \textbf{76.01}&   \textbf{71.78}&   \textbf{86.93}&   78.42&   \textbf{79.67}&   \textbf{79.20}&  77.52&   73.57&   66.06&   58.24&   68.85&  \textbf{68.66}&   85.06&   78.18&   79.28&   77.79\\
ch5.5&	 75.45&   76.49&   76.86&   76.27&   73.27&   87.62&   \textbf{77.80}&   81.31&   80.00&  \textbf{76.13}&   \textbf{72.75}&   66.17&   58.25&   \textbf{68.32}&  70.43&   85.36&   73.98&   79.51&   77.32\\\cdashline{1-20}[2.25pt/3pt]
db2&	 79.14&   79.52&   80.80&   79.82&   72.77&   86.98&   78.21&   81.21&   79.79&  78.23&   74.30&   \textbf{65.03}&   57.95&   68.88&  69.83&   85.96&   \textbf{69.29}&   81.04&   76.53\\\hline
	\end{tabular}}
	\end{center}
\end{table*}
\begin{figure}[bpt]
	\centering
	\includegraphics*[scale=0.6, viewport=23 4 416 320]{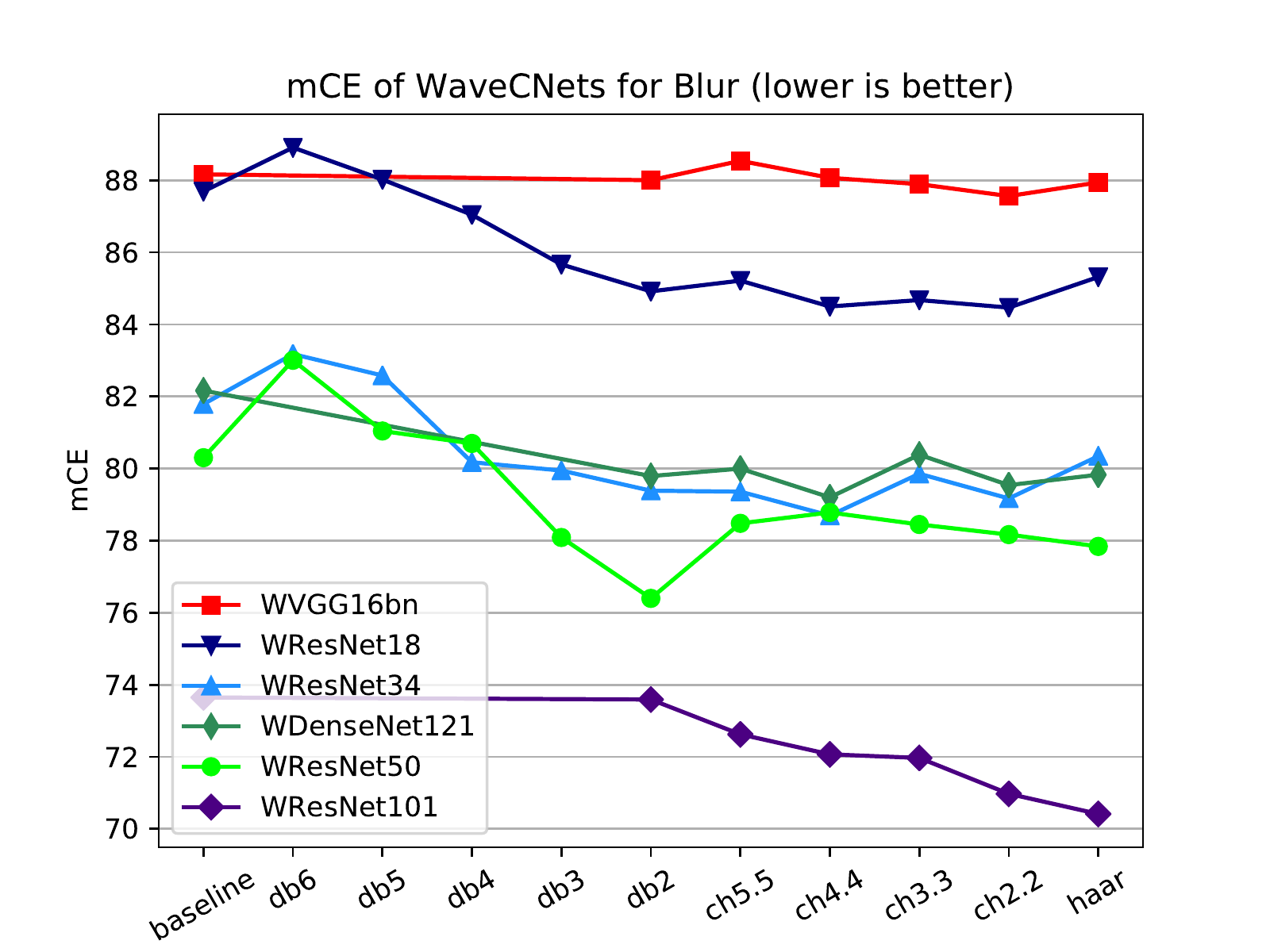}
	\caption{The blur mCE of WaveCNets.}
	\label{fig_mCE_blur}
\end{figure}
\begin{figure}[bpt]
	\centering
	\includegraphics*[scale=0.6, viewport=23 4 416 320]{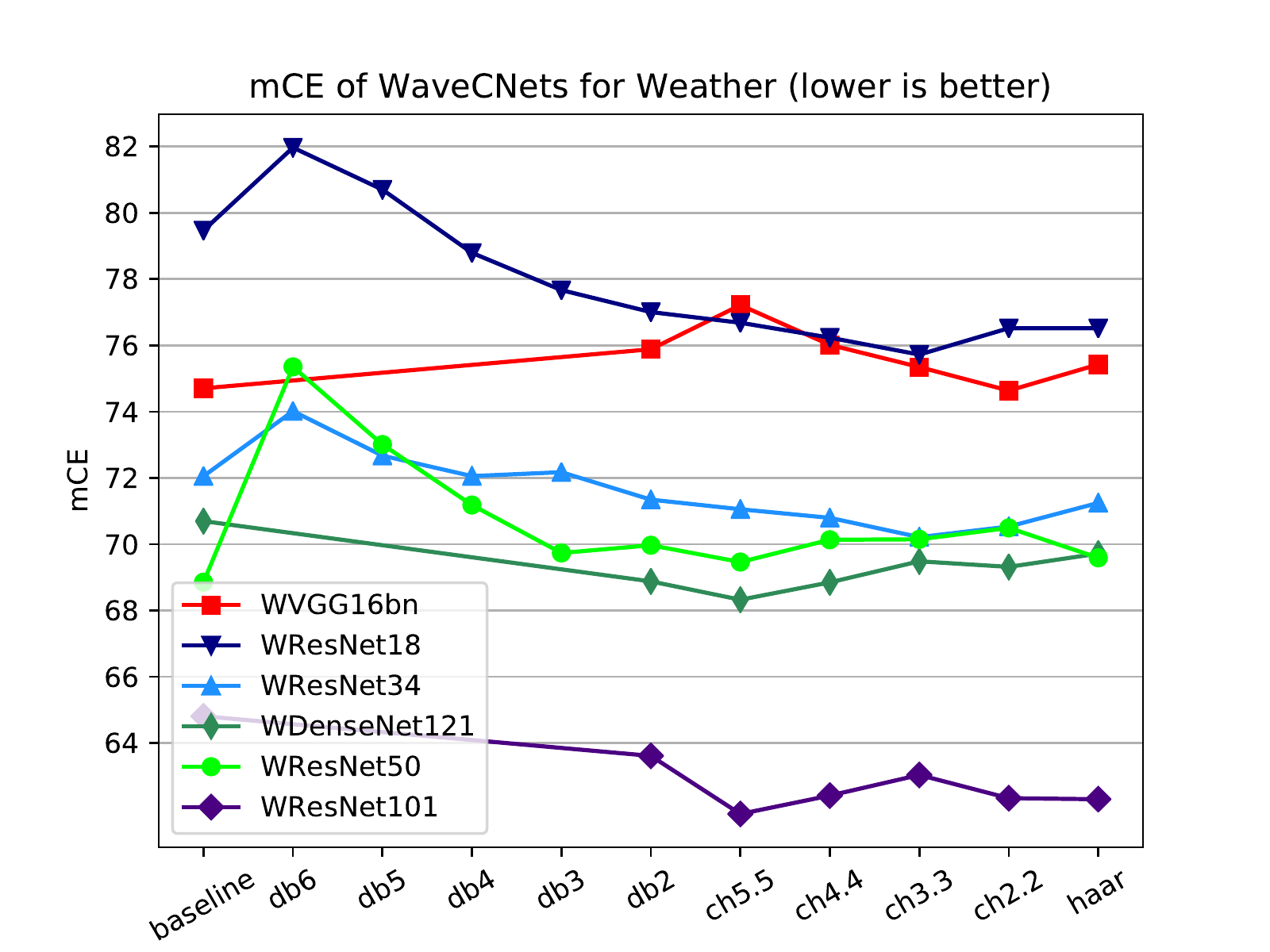}
	\caption{The weather mCE of WaveCNets.}
	\label{fig_mCE_weather}
\end{figure}
\begin{figure}[bpt]
	\centering
	\includegraphics*[scale=0.6, viewport=23 4 416 320]{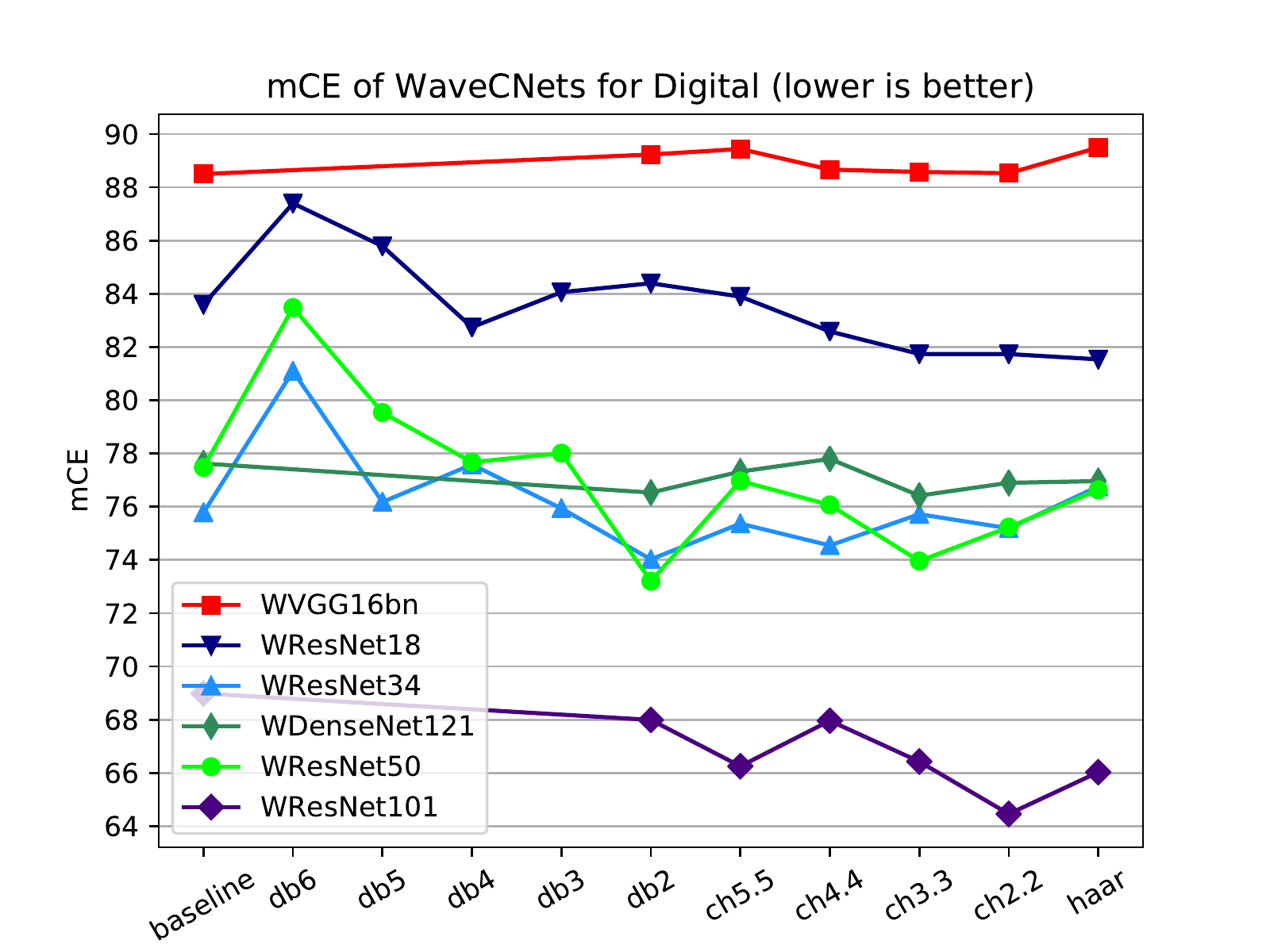}
	\caption{The digital mCE of WaveCNets.}
	\label{fig_mCE_digital}
\end{figure}

ImageNet-C is proposed in \cite{hendrycks2019benchmarking} to evaluate the robustness of a well-trained classifier to the common corruptions.
ImageNet-C contains various versions of ImageNet validation images produced with 15 visual corruptions with five severity levels.
We test WaveCNets on ImageNet-C, and compute the 15 CE values according to Eq. (\ref{eq_CE}).
Because the 15 corruptions are sourced form four categories,
i.e., noise (Gaussian noise, shot noise, impulse noise), blur (defocus blur, frosted glass blur, motion blur, zoom blur),
weather (snow, frost, fog, brightness), and digital (contrast, elastic, pixelate, JPEG-compression),
we compute four mCE according to Eq. (\ref{eq_mCE_noise}) and
\begin{align}
\label{eq_mCE_blur}
\text{mCE}_{\text{blur}}^f &= \dfrac{1}{4}\sum_{ c\in  \{\text{defocus},\ \text{glass}, \atop \quad\text{motion},\ \text{zoom}\}} \text{CE}_c^f,
\end{align}
\begin{align}
\label{eq_mCE_weather}
\text{mCE}_{\text{weather}}^f &= \dfrac{1}{4}\sum_{ c\in  \{\text{snow},\ \text{frost}, \atop \quad\text{fog},\ \text{bright}\}} \text{CE}_c^f,
\end{align}
\begin{align}
\label{eq_mCE_digital}
\text{mCE}_{\text{digital}}^f &= \dfrac{1}{4}\sum_{ c\in  \{\text{contrast},\ \text{elastic}, \atop \quad\text{pixel},\ \text{jpeg}\}} \text{CE}_c^f.
\end{align}

Table \ref{Tab_mCE_VGG16bn} - Table \ref{Tab_mCE_DenseNet121} present the detailed CE and mCE results for
WVGG16bn, WResNet18, WResNet34, WResNet50, WResNet101, WDenseNet121, respectively.
In these tables, the ``baseline'' corresponds to the results of original CNN architecture,
while ``haar'', ``chx.y'', and ``dbx'' correspond to that of WaveCNets with the wavelets.
For better illustration, Fig. \ref{fig_mCE_blur}, Fig. \ref{fig_mCE_weather}, and Fig. \ref{fig_mCE_digital}
show the mCE curves for the three corruptions, respectively.

\subsection*{D. Shift-invariance of WaveCNets}

In \cite{zhang2019making}, to evaluate the shift-invariance of a classifier $f$,
the author checks how often the classifier outputs the same predication, given the same image with two different shifts:
\begin{align}
\nonumber
\textbf{E}_{\textbf{X},h_1,w_1,h_0,w_0}\textbf{1}&\{\arg\max P_f(\text{shift}_{h_0,w_0}(\textbf{X})) \\
													&= \arg\max P_f(\text{shift}_{h_1,w_1}(\textbf{X}))\},
\end{align}
where $\textbf{E}$ denotes the expectation.
Using this criterion, we evaluate the shift-invariance of WaveCNets and the original CNNs.
Table \ref{Tab_shift_invariance} presents the results.
\begin{table}
\small
\caption{Shift-invariance of WaveCNets (higher is better).}
\label{Tab_shift_invariance}
\begin{center}
\setlength{\tabcolsep}{1.6mm}{
\begin{tabular}{r||c|cccc|c}\hline
\multirow{3}{*}{wavelet}	&\multicolumn{6}{c}{WaveCNet} \\\cline{2-7}
							&VGG & \multicolumn{4}{c|}{ResNet}&DenseNet  \\\cline{2-7}
							&16bn&18&34&50&101&121\\\hline
baseline&    89.24&     85.11&     87.56&      89.20&      89.81&     88.81\\\hline
haar    &    90.54&     86.43&     88.46&      89.93&      90.73&     89.12\\
ch2.2   &    91.03&     87.35&     89.02&      90.01&      91.06&     89.52\\
ch3.3   &    \textbf{91.32}&     \textbf{87.71}&     \textbf{89.11}&      \textbf{90.68}&      91.33&     \textbf{89.91}\\
ch4.4   &    90.68&     87.36&     89.04&      90.22&      \textbf{91.34}&     89.23\\
ch5.5   &    90.56&     86.97&     88.74&      89.94&      91.11&     89.33\\
db2     &    90.84&     86.96&     88.62&      89.69&      91.02&     89.05\\
db3     &         &     86.79&     88.50&      89.72&           &          \\
db4     &         &     86.44&     88.26&      89.47&           &          \\
db5     &         &     85.88&     87.96&      88.86&           &          \\
db6     &         &     84.84&     87.65&      87.87&           &          \\\hline
\end{tabular}}
\end{center}
\end{table}

\begin{figure*}[t]
	\centering
	\includegraphics*[scale=0.875, viewport=43 558 558 710]{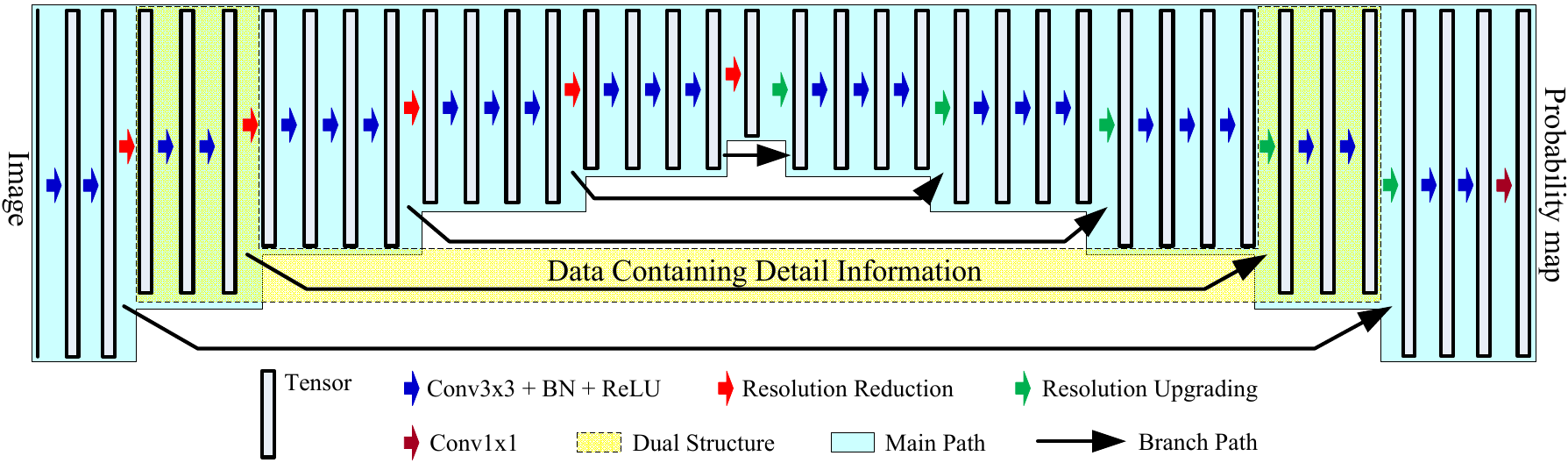}
	\caption{The encoder-decoder architectures.}\label{fig_encoder_decoder}
\end{figure*}
\subsection*{E. The architectures of SegNet and WaveUNets}
SegNet and WaveUNets adopt encoder-decoder architecture, as Fig. \ref{fig_encoder_decoder} shows.
The encoder consists of 13 convolutional layers
corresponding to the first 13 convolutional layers in the VGG16bn \cite{simonyan2014very}.
Their decoder contains the same number of convolutional layers with the encoder.
Every convolutional layer in the former and the corresponding one in the latter have the same number of channels,
except the first and the last one of the network.
In the encoder and decoder, a Batch Normalization (BN) and Rectified Linear Unit (ReLU) are implemented after every convolution.
A convolutional layer with kernel size of $1\times1$ converts the output of decoder into the predicted segmentation result,
as shown in Fig. \ref{fig_encoder_decoder}.
In Table \ref{Tab_network_configuration}, the first column shows the input size,
though these networks can process images with arbitrary size.
Every number in the table corresponds to a convolutional layer with BN and ReLU.
While the number in the column ``encoder'' is the number of the input channels of the convolution,
the number in the column ``decoder'' is the number of the output channels.
\begin{table}[!t]
	\caption{Deep network configurations.}
	\label{Tab_network_configuration}
	\begin{center}
	\begin{threeparttable}
		\setlength{\tabcolsep}{1.75mm}{
			\begin{tabular}{c||l|r}
				\hline
				\multirow{2}{*}{data size} &    \multicolumn{2}{c}{the number of channels}\\ \cline{2-3}
				                           & \multicolumn{1}{c|}{encoder} & \multicolumn{1}{c}{decoder}\\ \hline
				      $352\times480$       & 3, 64                    &                   64, 64 \\
				      $176\times240$       & 64, 128                  &                  128, 64 \\
				     ~~$88\times120$       & 128, 256, 256            &            256, 256, 128 \\
				       $44\times60$        & 256, 512, 512            &            512, 512, 256 \\
				       $22\times30$        & 512, 512, 512            &            512, 512, 512 \\ \hline
			\end{tabular}}
	\end{threeparttable}
	\end{center}
\end{table}

\subsection*{F. The amount of multiply-adds in 2D DWT/IDWT}
Given a 2D tensor $\textbf{X}$ with size of $M\times N$ and channel $C$,
the amount of multiply-adds used in 2D DWT inference is
\begin{align}
\label{eq_no_2D_DWT}
4C\left(M^2N+\dfrac{MN^2}{2} - \dfrac{3MN}{4}\right),
\end{align}
and the amount of multiply-adds used in 2D IDWT inference is
\begin{align}
\label{eq_no_2D_IDWT}
4C\left(MN^2+\dfrac{M^2N}{2} - \dfrac{3MN}{4}\right) + 3,
\end{align}
according to Eqs. (\ref{eq_DWT_2D_M_ll}) - Eqs. (\ref{eq_DWT_2D_M_hh}).

Table \ref{Tab_ratio_wavelet_convolution} presents
the ratios of wavelet related multiply-adds over the total operations for WaveCNets and WaveUNets,
when the input size is $3\times224\times224$.
We only count the amount of multiply-adds in $\text{DWT}_{ll}$ for WaveCNets.
\begin{table}
\scriptsize
\caption{Wavelet related operation ratios.}
\label{Tab_ratio_wavelet_convolution}
\begin{center}
\setlength{\tabcolsep}{0.75mm}{
\begin{tabular}{r||c|cccc|c|c}\hline
\multirow{3}{*}{}	&\multicolumn{6}{c|}{WaveCNet}& {WaveUNet}\\\cline{2-8}
							&VGG & \multicolumn{4}{c|}{ResNet}&DenseNet & VGG\\\cline{2-8}
							&16bn&18&34&50&101&121&16bn\\\hline
non-wavelet ($\times10^9$)&    $15.51$&     $1.82$& $3.67$ &$4.12$&$7.85$ & $2.88$&$30.77$\\\hline
wavelet ($\times10^9$)&    $1.43$&\multicolumn{4}{c|}{$0.22$}&$0.18$&$8.57$\\\hline
ratio ($\%$)   &    $8.44$&     $10.85$&    $5.69$& $5.11$& $2.75$&     $5.82$&$21.78$\\\hline
\end{tabular}}
\end{center}
\end{table}

\end{document}